\documentclass[10pt,twocolumn,letterpaper]{article}

\usepackage[pagenumbers]{cvpr} % To force page numbers, e.g. for an arXiv version

\usepackage{algorithm}
\usepackage{algpseudocode}
\usepackage{xcolor}
\usepackage{siunitx}
\usepackage{makecell}
\usepackage{amsmath}
\usepackage{caption}
\usepackage{varwidth}
\usepackage{wrapfig}
\usepackage{multirow}
\usepackage[T1]{fontenc}
\usepackage{tikz}
\usepackage{mathtools}
\usepackage{placeins}
\usepackage[accsupp]{axessibility}

\newcommand{\real}{\mathbb{R}}

\newcommand{\loss}{\mathcal{L}}

\definecolor{myred}{rgb}{0.9,0.0,0.0}
\definecolor{ours}{RGB}{120,190,255}
\definecolor{gtgray}{gray}{0.50}
\usepackage{pifont}

\newcommand{\greencheck}{\textcolor{green!80!black}{\ding{51}}}
\newcommand{\redx}{\textcolor{red}{\ding{55}}}

\newcommand{\highlightline}[1]{%
  \tikz[baseline] \node[anchor=base west, fill=ours!30, rounded corners=1pt, inner ysep=0pt, inner xsep=1pt] {\strut #1};%
}

\newcommand{\highlightlinetwo}[1]{%
  \tikz[baseline] \node[anchor=base west, fill=green!20, rounded corners=1pt, inner ysep=0pt, inner xsep=1pt] {\strut #1};%
}

\definecolor{cvprblue}{rgb}{0.21,0.49,0.74}
\usepackage[pagebackref,breaklinks,colorlinks,allcolors=cvprblue,linkcolor=myred]{hyperref}

\def\confName{CVPR}
\def\confYear{2026}

\title{Avatar Forcing: Real-Time Interactive Head Avatar Generation \\ for Natural Conversation}
\author{
Taekyung Ki$^{1,\star}$ ~~ Sangwon Jang$^{1,\star}$ ~~ Jaehyeong Jo$^1$ ~~ Jaehong Yoon$^2$ ~~ Sung Ju Hwang$^{1,3}$\\
$^1$KAIST \quad $^2$NTU Singapore \quad $^3$DeepAuto.ai\\
{\tt\small \{taekyung.ki, sangwon.jang, sungju.hwang\}\@kaist.ac.kr}\\
{\small \url{https://taekyungki.github.io/AvatarForcing}}
}

\begin{document}
% \maketitle
\twocolumn[{
\maketitle
\begin{center}
    \captionsetup{type=figure}
    \vspace*{-9mm}
    \includegraphics[width=\linewidth]{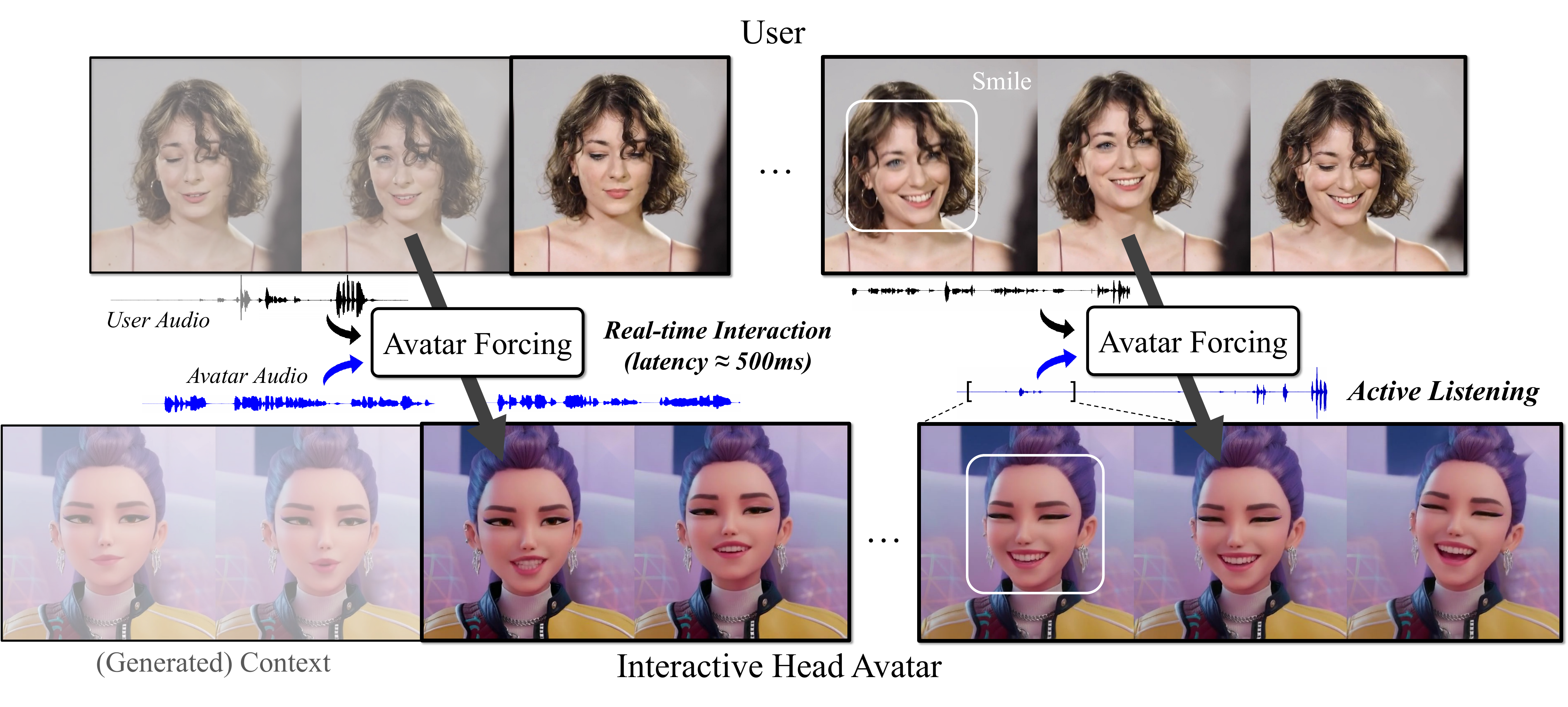}
    \vspace*{-8mm}
    \captionof{figure}{\textbf{Overview of Avatar Forcing}. It can generate a real-time interactive avatar video conditioned on user motion and audio, as well as avatar audio. The avatar naturally mirrors the user's expression, such as smiling when the user smiles, for more engaging interactions.  \label{fig:concept}
    }
\end{center}
}]

\def\thefootnote{\fnsymbol{footnote}}\footnotetext{\noindent\hspace*{-1em}$^\star$ Equal contribution.}\def\thefootnote{\arabic{footnote}}

%%%%%%%%%%%%%%%%%%%%%%%%%%%%%%%%%%%%%%%%%%%%%%%%%%%%%%%%%%%%%%%%%%%%%%%%%%%%%%%%%%%%%%%%%%%%%%%%

\begin{abstract}
Talking head generation creates lifelike avatars from static portraits for virtual communication and content creation.
However, current models do not yet convey the feeling of truly interactive communication, often generating one-way responses that lack emotional engagement. We identify two key challenges toward truly interactive avatars: generating motion in real-time under causal constraints and learning expressive, vibrant reactions without additional labeled data.
To address these challenges, we propose Avatar Forcing, a new framework for interactive head avatar generation that models real-time user-avatar interactions through diffusion forcing. This design allows the avatar to process real-time multimodal inputs, including the user’s audio and motion, with low latency for instant reactions to both verbal and non-verbal cues such as speech, nods, and laughter. Furthermore, we introduce a direct preference optimization method that leverages synthetic losing samples constructed by dropping user conditions, enabling label-free learning of expressive interaction. Experimental results demonstrate that our framework enables real-time interaction with low latency ($\approx$ \textit{500ms}), achieving 6.8$\times$ speedup compared to the baseline, and produces reactive and expressive avatar motion, which is preferred over 80\% against the baseline.
\end{abstract}    
\section{Introduction}\label{sec:intro}
Talking head generation animates static portrait images into lifelike avatars that can speak like humans. These systems are increasingly used to create virtual presenters, hosts, and educators that can substitute for real human presence in many scenarios. They also support customized avatars that users can interact with, for instance, chatting with their favorite characters~\citep{synthesia, hedra, pika}, offering a practical tool for content creation and visual communication.

However, existing avatar generation models fail to fully replicate the feeling of real face-to-face interaction. They primarily focus on generating audio-synchronized lip or natural head motions~\citep{vasa_1, float, emo, omnihuman, magicinfinite, talkingmachines, mocha, hunyuanportrait} to deliver information accurately and naturally, rather than engaging in interactive conversations. Such one-way interaction overlooks the bidirectional nature of real-life conversations, where the continuous exchange of verbal and non-verbal signals plays a crucial role in communication. For instance, active listening behaviors such as nodding or empathic responses encourage the speaker to continue, while expressive speaking behaviors such as smiling or making eye contact contribute to a more realistic and immersive conversation.

We identify two major challenges in generating truly interactive avatars. The first challenge is the real-time processing of users’ multimodal inputs. An interactive avatar system must continuously receive and respond to the user’s multimodal cues, such as speech, head motion, and facial expression,  which requires both low inference time and minimal latency. Existing methods~\citep{infp} achieve fast inference within a motion latent space~\citep{megaportrait} but have high latency because they need the full conversation context (e.g., over 3 seconds), including future frames. The model must wait for a sufficiently long audio segment before generating motion, causing a notable delay in user interaction, as illustrated in \cref{fig:blockwise_causal_dit}. This highlights the need for a causal motion generation framework that reacts immediately to live inputs.

The second challenge is learning expressive and vibrant interactive motion. Expressiveness in human interactions is inherently difficult to define or annotate, and the lack of well-curated data makes the natural interactive behaviors hard to model. For listening behaviors in particular, we observe that most of the training data are less expressive and low-variant, often exhibiting stiff posture (See \cref{fig:vis_speaker_listener_exp}). Moreover, unlike lip synchronization, which is strongly tied with the avatar audio and therefore relatively easy to learn, reacting appropriately to user cues can correspond to a wide range of plausible motions. This ambiguity greatly increases learning difficulty, often resulting in less diverse, stiff motions, particularly in response to non-verbal cues.

To address these challenges, we present a new interactive head avatar generation framework, \textbf{Avatar Forcing}, which models the causal interaction between the user and the avatar in a learned motion latent space~\citep{lia, float}. Inspired by recent diffusion-forcing-based interactive video generative models~\citep{self_forcing, causvid, genie3}, we employ causal diffusion forcing~\citep{diffusion_forcing, causvid, streamdit} to generate interactive head motion latents while continuously processing multimodal user inputs in real-time. Unlike the previous approach~\citep{infp} that requires the future conversation context (See~\cref{fig:blockwise_causal_dit}), Avatar Forcing causally generates interactive avatar videos by efficiently reusing past information through key-value (KV) caching. This design enables real-time reactive avatar generation with minimal latency.

Moreover, following the learning-from-losing paradigm of preference optimization methods~\citep{ouyang2022training, rafailov2023dpo, wallace2024diffusiondpo}, we propose a preference optimization method to enhance the interactiveness of the avatar motion. By synthesizing under-expressive motion latents via dropping the user signals and using them as less-preferred samples,
we achieve significantly improved expressiveness and reactiveness, without the need for additional human annotations for natural interaction. As a result, Avatar Forcing can produce more natural and engaging interactive videos as illustrated in~\cref{fig:concept}.

Extensive experiments demonstrate that Avatar Forcing supports real-time interaction with a latency of roughly 500ms. Moreover, the proposed preference optimization significantly improves motion reactiveness and richness, producing more natural and engaging videos. In human evaluations, our method is preferred over 80\% against the strongest baseline, demonstrating clear advantages in naturalness, responsiveness, and overall interaction quality. 

\section{Related Works} \label{sec:related}
\subsection{Talking Avatar Generation}
Talking avatar generation aims to generate a lifelike talking avatar video from a given reference image and an audio. Earlier methods in this field had focused on synthesizing accurate lip movement from the driving audio~\citep{wav2lip, makeittalk, stylelipsync, stylesync, keysync}. These approaches are extended to generate holistic head movements, including a rhythmical head motion~\citep{makeittalk, emo, float, vasa_1} and vivid facial expressions, such as eye blink and jaw movement~\citep{sadtalker, megaportrait, emoportraits, vasa_1, liveportrait}. SadTalker~\citep{sadtalker}, for instance, leverages 3D morphable models (3DMM)~\citep{bfm} as an intermediate representation for the non-verbal facial expression. EMO~\citep{emo} and its subsequent models~\citep{loopy, hallo3, echomimic} leverage the foundation image diffusion model (e.g., StableDiffusion~\citep{ldm}) for photorealistic portrait animation. 
Recent works~\citep{vasa_1, anitalker, float} introduce generative modeling techniques, such as diffusion models or flow matching into a learned motion space~\citep{megaportrait, lia}, achieving real-time head motion generation.

\subsection{Listening Avatar Generation}
Another line of works~\citep{l2l, rlhg, mfr_net, pch, customlistener, ditailistener} focus on generating realistic listening head motions, such as nodding or focusing. Generating such responsive motions is challenging because the relationship between the speaker and listener is inherently one-to-many~\citep{l2l, diffusion_listening}, and the cues are context-dependent and weakly supervised.
Hence, most of the works generate the \textit{personalized} listening motion~\citep{l2l, rlhg} or leverage explicit control signals, such as \textit{text instruction}~\citep{customlistener} and \textit{pose prior}~\citep{diffusion_listening}. 

\subsection{Dyadic Conversational Avatar Generation}
Recently, several studies investigated dyadic motion generation~\citep{dim, infp, arig}, which involves modeling the interactive behavior of two participants engaged in a conversational setting.
DIM~\citep{dim} quantizes dyadic motions into two discrete latent spaces~\citep{vqvae}, one for the verbal and one for the non-verbal. However, it requires a manual role-switching signal between the two spaces, resulting in discontinuous transitions between speaker and listener states. INFP~\citep{infp} addresses this limitation by introducing audio-conditioned verbal and non-verbal memory banks within a unified motion generator. 
However, this is ill-suited for real-time interactive generation as its bidirectional transformer~\citep{dit} requires access to the entire context of the conversation.
ARIG~\citep{arig} generates implicit 3D keypoints~\citep{liveportrait} as its motion representation, primarily focusing on facial expressions. Yet it struggles with temporal consistency and fails to produce holistic head motions.

In this work, we focus on modeling the interactive behavior of both participants continuously influencing each other through verbal and non-verbal signals. We generate holistic interactive motions, for instance, talking, head movement, listening, and focusing, using a diffusion forcing framework that promptly responds to the multimodal user signals. 

\section{Background} \label{sec:background}
\paragraph{Diffusion Forcing} Diffusion forcing~\citep{diffusion_forcing} stands out as an efficient sequential generative model that predicts \textit{next token} conditioned on \textit{the noisy past tokens}. Let $\mathbf{x}_1 = (x_1^{1}, x_1^2, \cdots, x_1^{N}) \in \real^{N \times 3 \times H \times W}$ denote a sequence of $N$ tokens sampled from data distribution $p_1$. Each token is corrupted with per-token independent noise levels $\boldsymbol{t}\coloneqq (t_1, t_2, \cdots, t_N) \in [0, 1]^{N}$, forming an independently noised sequence $\mathbf{x}_{\mathbf{t}} \coloneqq (x^1_{t_1}, x^2_{t_2}, \cdots, x^N_{t_N})$,
where $x^{n}_{t_n} = t_n x_1^{n} + (1-t_n)x_0^n$ and $\mathbf{x}_0 = (x_0^{n})_{n=1}^{N}\sim p_0$. Here, we follow the noise scheduler of flow matching~\citep{cfm}. The training objective of diffusion forcing is to regress the vector field $v_\theta$ toward the target vector field $v^n_{t_n} = x^n_1-x^n_0$:
\begin{equation}
	\loss_{DF}(\theta)  = \mathbb{E}_{n, t_n, x^n_{t_n}} \left[ \| v_{\theta}(x^n_{t_n}, t_n) -  v^n_{t_n}\| \right]. \label{eq:train_df}
\end{equation}

Diffusion forcing reformulates conventional teacher forcing in terms of diffusion models, allowing causal sequence generation with diffusion guidance~\citep{classifier_free} for controllable generation and flexible sampling procedure.

\paragraph{Direct Preference Optimization} Direct Preference Optimization (DPO)~\citep{rafailov2023dpo} aligns a model with human preferences without explicitly training a reward model. The training objective $\loss_{DPO}$ is formulated as follows:
\begin{equation}
    - \mathbb{E}_{c, x^w, x^l} \left[ \log \sigma \left( \beta \frac{\pi_{\theta} (x^w | c) }{\pi_{\text{ref}}(x^w | c)}  - \beta \frac{\pi_{\theta}(x^l | c)}{\pi_{\text{ref}}(x^l | c) } \right)\right], \label{eq:dpo}
\end{equation}
where $c$ is the condition, $x^w$ and $x^l$ denote \textit{preferred} and \textit{less preferred} samples, respectively, and $\pi_{\text{ref}}$ is a frozen reference model during the optimization, typically initialized by the pre-trained weight of $\pi_\theta$. Here, $\sigma(\cdot)$ is the sigmoid function, and $\beta$ is the deviation parameter. 
DiffusionDPO~\citep{wallace2024diffusiondpo} extends DPO to diffusion models by reformulating the objective with diffusion likelihoods, enabling preference optimization using the evidence lower bound.

\section{Avatar Forcing}\label{sec:method}
\begin{figure*}[t]
    \centering
    \vspace{-0.1in}
    \includegraphics[width=\linewidth]{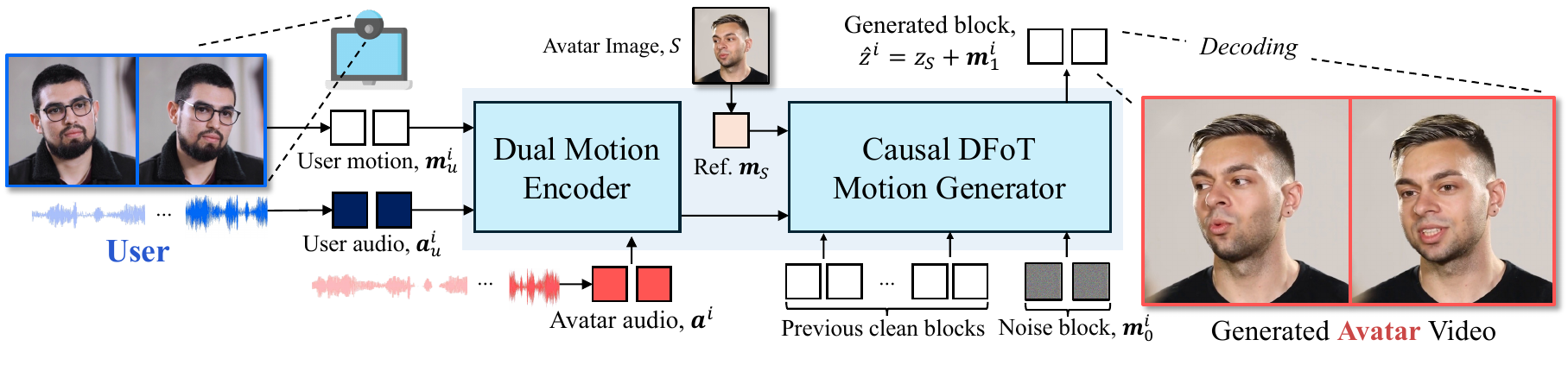}
    \vspace*{-8mm}
    \caption{\textbf{Overall architecture of Avatar Forcing}. We encode the use motion and audio, as well as avatar audio into a unified condition by Dual Motion Encoder. Causal Motion Generator infer the motion latent block of the avatar, which are then decoded into an avatar video.} \label{fig:architecture1}
    \vspace*{-2mm}
\end{figure*}

In this work, we present Avatar Forcing, which generates a video of a real-time interactive head avatar, conditioned on the avatar audio and the multimodal signals of the user. We provide an overview of our framework in~\cref{fig:architecture1}.

In \cref{sec:df_motion_sample}, we present our framework for achieving real-time interactive head avatar generation based on causal diffusion forcing in the motion latent space. This consists of two key steps: encoding multimodal user and avatar signals, and causal inference of avatar motion.
In \cref{sec:dpo}, we introduce a preference optimization method for the interactive motion generation, which enhances expressive interaction without the need for additional human labels.

%%%%%%%%%%%%%%%%%%%%%%%%%%%%%%%%%%%%%%%%%%%%%%%%%%%%%%%%%%%%%%%%%%%%%%%%%%%%%%%%%%%%%%%%%%%%%%%%
%%%%%%%%%%%%%%%%%%%%%%%%%%%%%%%%%%%%%%%%%%%%%%%%%%%%%%%%%%%%%%%%%%%%%%%%%%%%%%%%%%%%%%%%%%%%%%%%

%%%%%%%%%%%%%%%%%%%%%%%%%%%%%%%%%%%%%%%%%%%%%%%%%%%%%%%%%%%%%%%%%%%%%%%%%%%%%%%%%%%%%%%%%%%%%%%%
\begin{figure}[t]
    \centering
    \vspace{-0.05in}
    \includegraphics[width=0.49\textwidth]{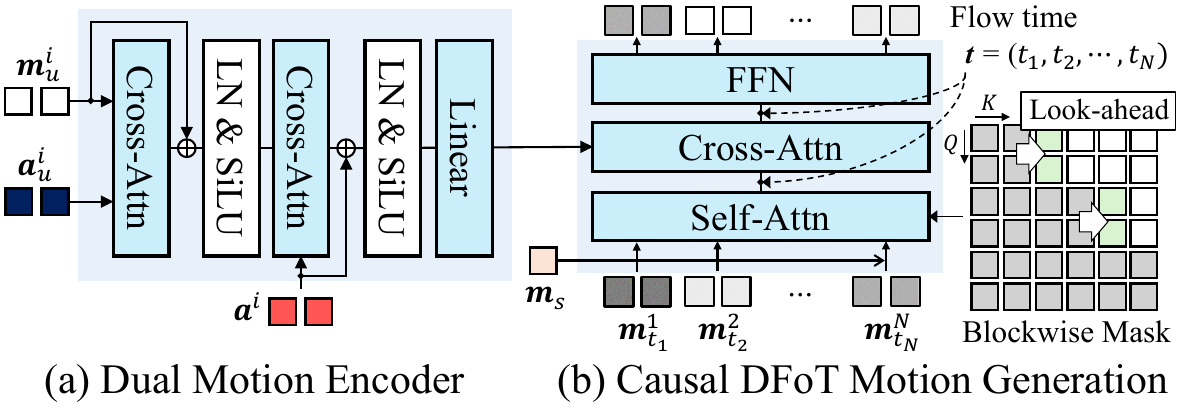}
    \vspace*{-6mm}
    \caption{\textbf{Architecture of} $v_{\theta}$. The look-ahead causal attention mask enables a smooth transition across the blocks.}
    \label{fig:architecture2}
\vspace{-0.2in}
\end{figure}
%%%%%%%%%%%%%%%%%%%%%%%%%%%%%%%%%%%%%%%%%%%%%%%%%%%%%%%%%%%%%%%%%%%%%%%%%%%%%%%%%%%%%%%%%%%%%%%%

\subsection{Diffusion Forcing for Real-Time Interaction} \label{sec:df_motion_sample}
\paragraph{Motion Latent Encoding} 
For motion encoding, we employ the motion latent auto-encoder from \citet{float}. The latent auto-encoder maps an input image $S\in\real^{3 \times\! H \!\times\! W}$ to a latent $z \in \real^d$ whose identity and motion are decomposable:
\begin{equation}
    z = z_{S} + \mathbf{m}_S \in \real^d,
\end{equation}
where $z_{S}\in\real^d$ and $\mathbf{m}_S\in\real^d$ are the identity and motion latents, respectively. The identity latent $z_S$ encodes identity representation of the avatar image $S$ (e.g., appearance), while the motion latent $\mathbf{m}_S$ encodes rich verbal and non-verbal features (e.g., facial expression, head motion). We use this latent representation to capture holistic head motion and fine-grained facial expression, which are crucial for realistic head avatar generation.
We provide more details on the auto-encoder in Appendix~\ref{sec:apx_architecture}.

\paragraph{Interactive Motion Generation}
In Avatar Forcing, the motion latents are generated by conditioning on multimodal user signals and avatar audio. This can be formulated as an autoregressive model as follows:
\begin{equation}
    \vspace*{-1mm}
    p_{\theta}( \mathbf{m}^{1:N}) = \prod_{i=1}^{N} p_{\theta}(\mathbf{m}^i | \mathbf{m}^{<i}, \mathbf{c}^{\leq i}), \label{eq:task}
    \vspace*{-1mm}
\end{equation}
where each motion latent $\mathbf{m}^i$ is predicted from past motion latents and the condition triplet $\mathbf{c}^{i} = (\mathbf{a}_u^{i}, \mathbf{m}_u^i, \mathbf{a}^i)$ consisting of user audio $\mathbf{a}_u^{i}$, user motion $\mathbf{m}_u^i$, and avatar audio $\mathbf{a}^i$.

Based on this formulation, we introduce a diffusion forcing-based causal motion generator operating in the motion latent space, which is modeled using a vector field model $v_\theta$. As illustrated in~\cref{fig:architecture2}, the model $v_\theta$ comprises two main components: Dual Motion Encoder and Causal DFoT Motion Generation.

The goal of the Dual Motion Encoder is to capture the bidirectional relationship between multimodal user signals and avatar audio, and to encode them into a unified condition. As illustrated in~\cref{fig:architecture2}(a), the encoder first takes the user signals ($\mathbf{m}^{i}_u$ and $\mathbf{a}^{i}_u$) and aligns them through a cross-attention layer, which captures the holistic user motion. This representation is then integrated with the avatar audio using another cross-attention layer, which learns the causal relation between the user and the avatar, producing a unified user-avatar-condition. 
In \cref{sec:experiments:ablation}, we empirically validate the importance of using user motion $\mathbf{m}^{i}_u$ for generating an interactive avatar.

For the causal motion generator, we adopt the diffusion forcing transformer (DFoT)~\citep{historydiffusion} with a blockwise causal structure~\citep{causvid, ar_wo_vq}. The latent frames are divided into blocks to capture local bidirectional dependencies within each block while maintaining causal dependencies across different blocks. 
For each block, we assign a shared noise timestep to all frames and apply an attention mask that prevents the current block from attending to any future blocks.

\begin{figure}[t]
    \centering
    \vspace{-0.05in}
    \includegraphics[width=0.48\textwidth]{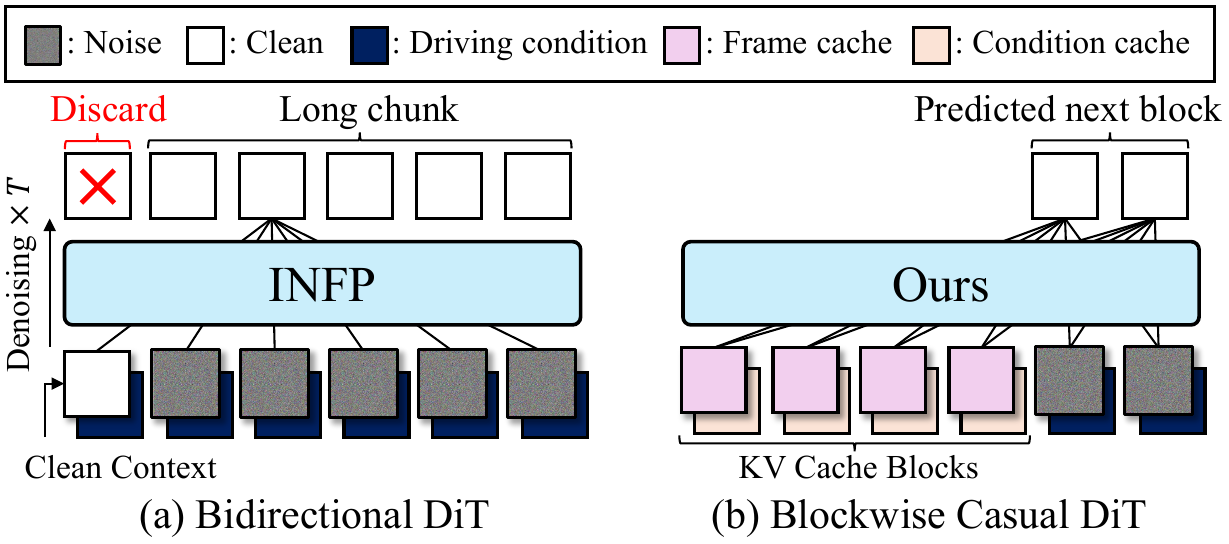}
    \vspace*{-7mm}
    \caption{\textbf{Architectural comparison between bidirectional and causal structure}. (a) Bidirectional DiT used in INFP~\citep{infp} requires access to the entire temporal window for motion generation. (b) Our blockwise causal DFoT predicts the next block without using future context and supports KV caching.
} \label{fig:blockwise_causal_dit}
\vspace{-0.15in}
\end{figure}

In \cref{fig:blockwise_causal_dit}, we compare our motion generator architecture with the standard bidirectional DiT architecture used in INFP~\citep{infp}. Unlike INFP, which requires the full temporal context, our diffusion forcing allows stepwise motion generation under causal constraints and user-avatar interaction with low latency. 

%%%%%%%%%%%%%%%%%%%%%%%%%%%%%%%%%%%%%%%%%%%%%%%%%%%%%%%%%%%%%%%%%%%%%%%%%%%%%%%%%%%%%%%%%%%%%%%%
%%%%%%%%%%%%%%%%%%%%%%%%%%%%%%%%%%%%%%%%%%%%%%%%%%%%%%%%%%%%%%%%%%%%%%%%%%%%%%%%%%%%%%%%%%%%%%%%

However, we observe that the strict causal mask in the blockwise causal attention fails to ensure a smooth temporal transition across blocks.
To address this issue, we introduce \emph{look-ahead} in the causal mask, which allows each block to attend to a limited number of future frames while preserving overall causality.
We define this blockwise look-ahead causal mask $M$ as follows (illustrated in~\cref{fig:architecture2}(b), right):
\begin{align}
    M_{i,j} = 1 ~~\text{if}~~ \lfloor j / B \rfloor \leq \lfloor i / B \rfloor + l ~~\text{else}~~ 0, \label{eq:lookahead_mask}
\end{align}
where $i$ and $j$ are the frame indices, $B$ denotes the block size, and $l$ is the look-ahead frame size. With look-ahead, we effectively mitigate severe per-frame motion jittering observed in the simple blockwise causal structure, which we provide video examples in the supplementary materials.

Based on the blockwise causal structure, the motion generator takes a noisy latent block as input, which is concatenated with the avatar motion latent $\mathbf{m}_S$ serving as the reference motion.
We use a cross-attention layer to condition on the unified condition from the Dual Motion Encoder. 
Additionally, we employ a sliding-window attention mask of size $2l$ along the time axis for temporal smooth conditioning. We provide the more details on $v_{\theta}$ in Appendix~\ref{sec:apx_architecture}.

%%%%%%%%%%%%%%%%%%%%%%%%%%%%%%%%%%%%%%%%%%%%%%%%%%%%%%%%%%%%%%%%%%%%%%%%%%%%%%%%%%%%%%%%%%%%%%%%
\begin{figure}[t]
\vspace{-0.15in}
\centering
\begin{minipage}{1.0\linewidth}
\begin{algorithm}[H]
\small 
\caption{\label{algo:sampling} Motion inference with KV caching} 
\begin{algorithmic}[1]
\Require ODE timesteps $\{t_n\}^T_{n=0}$, motion generator $v_{\theta}$, video length $N$, block size $B$, max cache size $M$, user inputs $(\mathbf{a}_{u},\mathbf{m}_{u})$, avatar audio $\mathbf{a}^i$, latent-to-frame decoder $\texttt{Dec}$, and id latent $z_S$. 
\State \textbf{Divide} the frames into $L = \lceil N / B \rceil$ blocks
\State \textbf{Initialize} $\mathbf{KV}, \mathbf{cKV}$ $\leftarrow [~], [~]$ $\rhd$ Frame \& condition caches
\For{$i = 1$ to $L$}
    \State \textbf{Sample} Noise block $\mathbf{m}_{t_{0}}^{i} \sim \mathcal{N}(0, I)$, \hfill $\rhd$ $\mathbf{m}_{t_{0}}^{i}\in\real^{B \times d}$
    \State \textbf{Acquire} \highlightline{User inputs ($\mathbf{a}^i_{u}$, $\mathbf{m}_{u}^i$)
    and avatar audio $\mathbf{a}^i$}.
    \State \textbf{Set} $\mathbf{c}^i$ $\leftarrow$ $(\mathbf{a}^i_{u}, \mathbf{m}_{u}^i, \mathbf{a}^i)$ \hfill $\rhd$ Condition triplet
    \For{$j = 0$ to $T$}
        \State \textbf{Solve ODE:} $\mathbf{m}_{t_{j+1}}^{i} \leftarrow v_{\theta}(\mathbf{m}_{t_j}^{i}, t_j; \mathbf{c}^i, \mathbf{KV}, \mathbf{cKV})$
    \EndFor
    \State \textbf{Decode} \& \textbf{Return}~~ \highlightline{$\mathbf{x}_1^i$ $\leftarrow$ $\texttt{Dec}(z_S, \mathbf{m}_1^i)$~$\in$~$\real^{B \times 3 \times H \times W}$}
    \State \textbf{Update caches}~~ $\mathbf{kv}_i$, $\mathbf{ckv}_i$ $\leftarrow$ $v_{\theta}(z_{1}^{i}, 1; \mathbf{c}^i, \mathbf{KV}, \mathbf{cKV})$    
    \If{$|\mathbf{KV}|$~$=$~$|\mathbf{cKV}|$~$=$~$M$}
    \State $\mathbf{KV}{\texttt{.pop(0)}}$~and~  $\mathbf{cKV}{\texttt{.pop(0)}}$
    \EndIf
    \State $\mathbf{KV}${\texttt{.append($\mathbf{kv}_i$)}}~and~ $\mathbf{cKV}${\texttt{.append($\mathbf{ckv}_i$)}}
\EndFor
\end{algorithmic}
\end{algorithm}
\end{minipage}
\vspace{-0.3in}
\end{figure}
%%%%%%%%%%%%%%%%%%%%%%%%%%%%%%%%%%%%%%%%%%%%%%%%%%%%%%%%%%%%%%%%%%%%%%%%%%%%%%%%%%%%%%%%%%%%%%%%

\paragraph{Training and Inference}
To train the vector field model $v_{\theta}$, we formulate the diffusion forcing training objective in Eq.~\eqref{eq:train_df} into a motion latent generation objective as follows:
\begin{equation}
	\loss_{DF}(\theta) = \mathbb{E}_{n, t_n, \mathbf{m}^n_{t_n}}\! \| v_{\theta}(\mathbf{m}^n_{t_n}, t_n, \mathbf{c}^n) \!-\! (\mathbf{m}_1^n \!-\! \mathbf{m}_0^n)\|,
    \label{eq:train_ours}
\end{equation}
where $n$~$\in$~$[1,N]$ denotes the frame index, $t_n$~$\in$~$[0,1]$ is the per-frame noise timestep, $\mathbf{c}^n = (\mathbf{a}_u^{i}, \mathbf{m}_u^i, \mathbf{a}^i)$ is the user-avatar condition triplet, $\mathbf{m}_{t_n}^n$~$\in$~$\real^d$ is the noisy motion latent. For simplicity, we omit the reference motion condition $\mathbf{m}_S$ in Eq.~\eqref{eq:train_ours}.

With the trained model, we can generate the interactive avatar motion given user inputs and avatar audio in an autoregressive rollout. We provide the pseudo code for motion sampling in \cref{algo:sampling}, which adopts rolling KV cache in a blockwise manner~\citep{causvid, self_forcing}. Further details on inference are provided in Appendix~\ref{sec:apx_exp_detail}.

\subsection{Enhancing Motion Expressiveness} \label{sec:dpo}
While our model enables real-time motion generation, we observe that it struggles to produce expressive motions crucial for natural human conversation. In contrast to achieving an accurate lip-sync that has an almost one-to-one match with the avatar audio, appropriate reaction to the user's motion and audio is highly ambiguous, as there is no single \emph{correct} response~\citep{l2l, diffusion_listening, customlistener}. 

\begin{figure}[t]
\centering
    \vspace{-0.05in}
    \includegraphics[width=\linewidth]{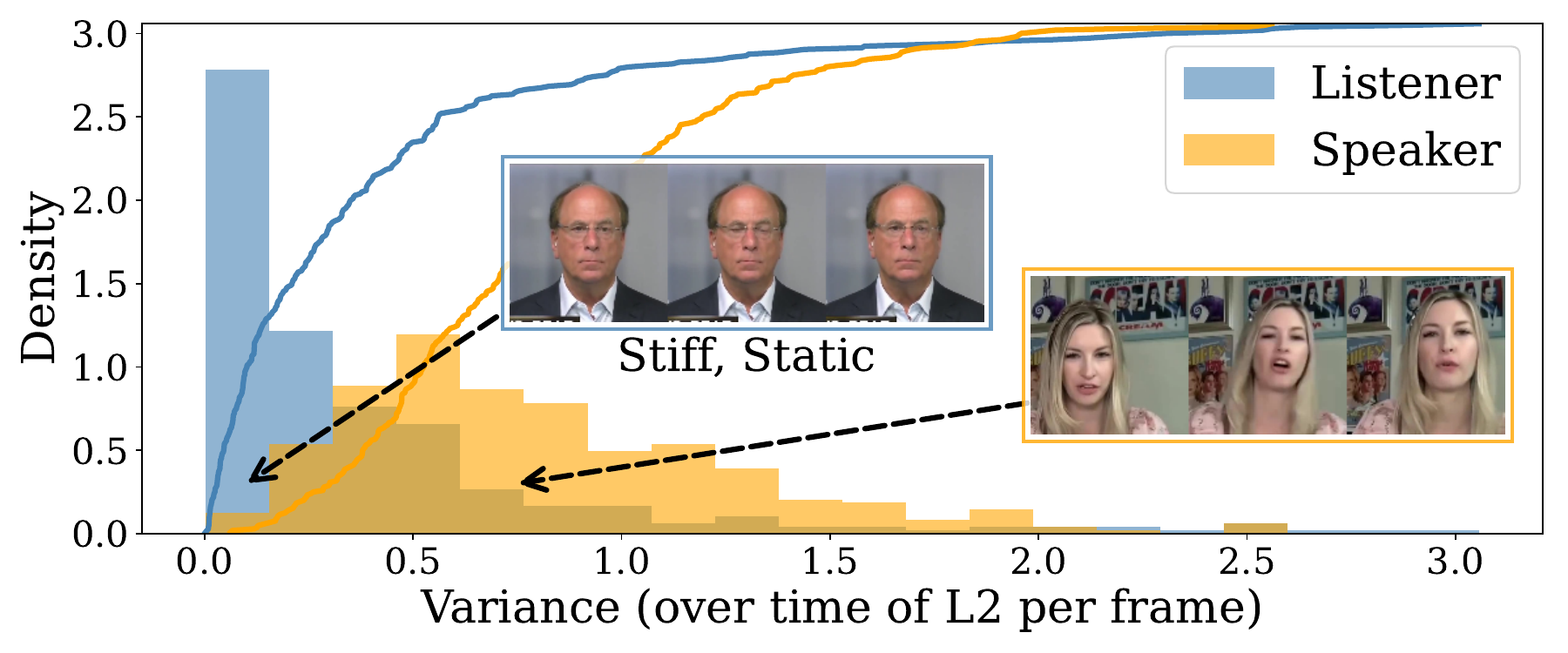}
    \vspace{-0.3in}
    \caption{\textbf{Variance visualization} of the L2-norm of 3DMM expressions~\citep{spectre} for the speaker and listener on ViCo~\citep{rlhg} dataset. Higher variance indicates higher expressiveness.}\label{fig:vis_speaker_listener_exp}
\vspace{-0.15in}
\end{figure}

Moreover, as shown in \cref{fig:vis_speaker_listener_exp}, existing listening datasets generally exhibit lower motion expressiveness compared to speaking datasets, which leads models to learn passive and non-expressive listening behaviors.
While prior work attempted to address it with personalized motion~\citep{l2l} or external text instructions~\citep{customlistener}, these methods do not address the core challenge of learning the true interactive behavior for natural interaction.

To generate vibrant and expressive reactions, we formulate this as an alignment problem and apply the Reinforcement Learning from Human Feedback (RLHF)~\citep{ouyang2022training} approach. 
The main challenge lies in defining an explicit reward for the avatar’s interactive behavior, which must account for the interplay between avatar audio, user audio, user motion, and the generated video, which are difficult to evaluate even for humans.

We adapted a reward-free method, Direct Preference Optimization (DPO)~\citep{rafailov2023dpo, wallace2024diffusiondpo}, for fine-tuning our motion generation based on diffusion forcing. Specifically, we construct preference pairs of motion latents $(\mathbf{m}^w, \mathbf{m}^l)$ as follows:
\begin{itemize}[leftmargin=1.5em]
\item \textbf{Preferred sample} $\mathbf{m}^w$: the motion latent from the ground-truth video, exhibiting expressive and contextually appropriate responses.
\item \textbf{Less preferred sample} $\mathbf{m}^l$: the motion latent generated by a separately trained \textit{talking avatar model}~\citep{float}, conditioned solely on the avatar audio.
\end{itemize}
This paired design yields a clear signal for enhancing expressiveness, emphasizing active listening and reactive motion while leaving other aspects, such as lip sync or speech-driven motion, unchanged.

Building on the original DPO objective $\loss_{\text{DPO}}$ (Eq.~\eqref{eq:dpo}) with our constructed pairs $(\mathbf{m}^w, \mathbf{m}^l)$, we fine-tune our diffusion forcing model $v_{\theta}$ with the following objective:

\begin{equation}
    \loss_{ft}(\theta) = \loss_{DF}(\theta) + \lambda\loss_{DPO}(\theta), \label{eq:total_objective}
\end{equation}
where $\lambda$ is a balancing coefficient. As a result, our model achieves efficient preference alignment without requiring a dedicated reward model, which we validate in \cref{sec:experiments:ablation}. We provide the detailed formulation of $\loss_{\text{DPO}}$ in Appendix~\ref{sec:apx_dpo}.

%%%%%%%%%%%%%%%%%%%%%%%%%%%%%%%%%%%%%%%%%%%%%%%%%%%%%%%%%%%%%%%%%%%%%%%%%%%%%%%%%%%%%%%%%%%%%%%%
%%%%%%%%%%%%%%%%%%%%%%%%%%%%%%%%%%%%%%%%%%%%%%%%%%%%%%%%%%%%%%%%%%%%%%%%%%%%%%%%%%%%%%%%%%%%%%%%
%%%%%%%%%%%%%%%%%%%%%%%%%%%%%%%%%%%%%%%%%%%%%%%%%%%%%%%%%%%%%%%%%%%%%%%%%%%%%%%%%%%%%%%%%%%%%%%%
\begin{figure*}[t]
\vspace{-0.05in}
\captionof{table}{\textbf{Quantitative comparison results} on RealTalk~\citep{realtalk}. Best results highlighted in \textbf{bold}. $^{*}$ denotes the reproduced version that is publicly unavailable. We also report the results from a non-interactive talking head model~\citep{float}, shown in gray, for reference.
} \label{tab:quantitative_realtalk}
\centering
\begin{minipage}{1\linewidth}
    \vspace*{-2mm}
    \resizebox{1.0\textwidth}{!}{
        \renewcommand{\arraystretch}{0.94}
        \renewcommand{\tabcolsep}{4pt}
        \begin{tabular}{l c cc c c c  c  c c  c c}
        \toprule
         & \multicolumn{2}{c}{Interaction} & \multicolumn{2}{c}{Reactiveness} & \multicolumn{2}{c}{Motion Richness} & \multicolumn{3}{c}{Visual Quality} & \multicolumn{2}{c}{Lip Synchronization} \\
        \cmidrule(l{2pt}r{2pt}){2-3}
        \cmidrule(l{2pt}r{2pt}){4-5}
        \cmidrule(l{2pt}r{2pt}){6-7}
        \cmidrule(l{2pt}r{2pt}){8-10}
        \cmidrule(l{2pt}r{2pt}){11-12}
        Method & User Input & Latency $\downarrow$ & rPCC-Exp $\downarrow$ & rPCC-Pose $\downarrow$ & SID $\uparrow$ & Var $\uparrow$ & FID $\downarrow$ & FVD $\downarrow$ & CSIM $\uparrow$ & LSE-D $\downarrow$ & LSE-C $\uparrow$ \\
        \midrule
        \textcolor{gray}{FLOAT}~\citep{float}           & \redx & \textcolor{gray}{2.4s} & 
        \textcolor{gray}{0.054} & \textcolor{gray}{0.182} & \textcolor{gray}{2.785} & \textcolor{gray}{2.778} & \textcolor{gray}{81.297} & \textcolor{gray}{438.817} & \textcolor{gray}{0.845}  & \textcolor{gray}{8.135} & \textcolor{gray}{6.361} \\
        % \midrule
        INFP$^{*}$~\citep{infp}                   & \greencheck & 3.4s & 0.035 & 0.064 & 2.343 & 1.638 & 24.551 & \textbf{159.000} & \textbf{0.867} & \textbf{8.027} & 6.536 \\
        \textbf{Ours}                & \greencheck & \textbf{0.5s} & \textbf{0.003} & \textbf{0.036} & \textbf{2.442} & \textbf{1.734}  & \textbf{24.328} & 170.874 & 0.833  &  8.060 & \textbf{6.723}\\
        \midrule        
        GT       & N/A & N/A & 0.000 & 0.000 & 3.972 & 1.658 & N/A & N/A & 0.796 & 7.790 & 6.940 \\
        \bottomrule
        \end{tabular}
        }
\end{minipage}
\centering
\begin{minipage}{1.0\textwidth}
    \centering
    \vspace*{0.05in}
    \caption{\textbf{Human preference study} on  interactive avatar generation models, comparing Avatar Forcing and INFP$^{*}$.} 
    \vspace*{-0.15in}
    \includegraphics[width=0.98\textwidth]{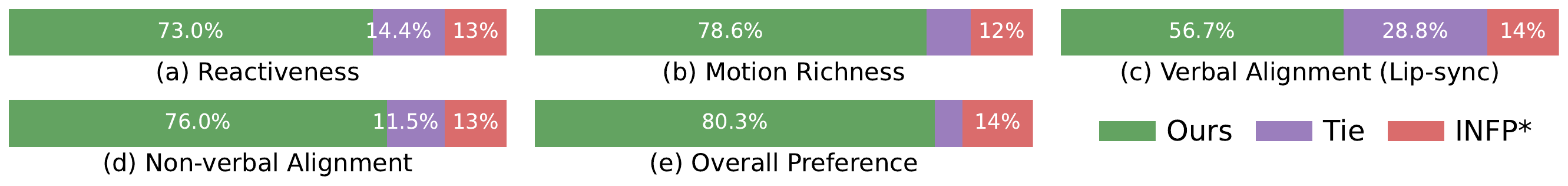}
    \label{fig:human_eval}
    \vspace*{-0.1in}
\end{minipage}
\end{figure*}
%%%%%%%%%%%%%%%%%%%%%%%%%%%%%%%%%%%%%%%%%%%%%%%%%%%%%%%%%%%%%%%%%%%%%%%%%%%%%%%%%%%%%%%%%%%%%%%%

%%%%%%%%%%%%%%%%%%%%%%%%%%%%%%%%%%%%%%%%%%%%%%%%%%%%%%%%%%%%%%%%%%%%%%%%%%%%%%%%%%%%%%%%%%%%%%%%
\begin{figure*}[t]
\begin{minipage}[t]{0.49\linewidth}
    \captionof{table}{\textbf{Comparison with talking head generation models} on the HDTF~\citep{hdtf} dataset. Second-best results are underlined.
    } \label{tab:compare_talking_hdtf}
    \vspace*{-1mm}
    \centering
    \resizebox{1.0\linewidth}{!}{
    \renewcommand{\arraystretch}{1.1}
    \renewcommand{\tabcolsep}{5pt}
    \begin{tabular}{lccccc}
    \toprule
     & \multicolumn{3}{c}{Visual Quality} & \multicolumn{2}{c}{Lip Synchronization} \\
            \cmidrule(l{2pt}r{2pt}){2-4}
            \cmidrule(l{2pt}r{2pt}){5-6}
                             Method & FID $\downarrow$         & FVD $\downarrow$          & CSIM $\uparrow$        & LSE-D $\downarrow$         & LSE-C $\uparrow$        \\
                             \midrule
    SadTalker \cite{sadtalker}     &   64.744              &  342.996               &  0.697             &  8.046              & 7.171           \\
    Hallo3  \cite{hallo3}          &   32.794              &  184.341               &  0.865             &  8.498              & 7.487           \\
    FLOAT \cite{float}             &   \underline{25.110}  &  \underline{167.463}   &  \textbf{0.881}    &  \textbf{7.553}     & \textbf{8.006}  \\
    INFP*~\citep{infp}             &   27.155              &  187.977               &  0.840             &  7.810              & 7.325          \\
    \midrule
    \textbf{Ours}                           &   \textbf{20.332}     &  \textbf{149.798}     &  \underline{0.870}  &  \underline{7.700}       &   \underline{7.560}       \\
    \bottomrule
    \end{tabular}}
\end{minipage}
\hfill
\begin{minipage}[t]{0.49\linewidth}
    \captionof{table}{\textbf{Comparison with listening head generation models} on the ViCo~\citep{rlhg} dataset. $\dagger$~denotes results inherited from DIM~\citep{dim}.} \label{tab:compare_listening_vico}
    \vspace*{-1mm}
    \centering
    \resizebox{1.0\linewidth}{!}{
    \renewcommand{\arraystretch}{1.07}
    \renewcommand{\tabcolsep}{4pt}
    \begin{tabular}{l cc cc cc cc}
    \toprule
     & \multicolumn{2}{c}{FD $\downarrow$}  & \multicolumn{2}{c}{rPCC $\downarrow$}  & \multicolumn{2}{c}{SID $\uparrow$} & \multicolumn{2}{c}{Var $\uparrow$} \\
    \cmidrule(l{2pt}r{2pt}){2-3}
    \cmidrule(l{2pt}r{2pt}){4-5}
    \cmidrule(l{2pt}r{2pt}){6-7}
    \cmidrule(l{2pt}r{2pt}){8-9}
            Method  & Exp  & Pose  & Exp  & Pose  &  Exp  &  Pose  &  Exp  &  Pose       \\
            \midrule
    RLHG$^\dagger$~\citep{rlhg}   &   39.02          &  0.07            &  0.08         &  0.02           & 3.62    &   3.17     &  1.52   &  0.02           \\
    L2L$^\dagger$~\citep{l2l}     &   33.93          &  0.06            &  0.06         &  0.08           & 2.77    &   2.66     &  0.83   &  0.02           \\
    DIM$^\dagger$~\citep{dim}     &   23.88          &  0.06            &  0.06         &  0.03           & \textbf{3.71}    &   2.35     &  1.53   &  0.02           \\
    INFP*~\citep{infp}  &   17.52              &  0.07               &  \textbf{0.01}            &  0.07             & 2.19 & \textbf{3.20}      &  2.10 &  0.03           \\
    \midrule
    \textbf{Ours}      &   \textbf{16.64}              &  \textbf{0.05}               &  \textbf{0.01}            &  \textbf{0.01}             & 3.12      &   3.00   &  \textbf{2.80} &  \textbf{0.03}           \\
    \bottomrule
    \end{tabular}
}
\vspace{-0.2in}
\end{minipage}
\vspace{-0.1in}
\end{figure*}

%%%%%%%%%%%%%%%%%%%%%%%%%%%%%%%%%%%%%%%%%%%%%%%%%%%%%%%%%%%%%%%%%%%%%%%%%%%%%%%%%%%%%%%%%%%%%%%%
\section{Experiments}\label{sec:expriments}
\subsection{Dataset and Preprocessing}
We use dyadic conversation video datasets: RealTalk~\citep{realtalk} and ViCo~\citep{rlhg}. We first detect scene changes using PySceneDetect~\citep{pyscenedetect} to split the videos into individual clips. We then detect and track each face using Face-Alignment~\citep{face_alignment}, crop and resize it to $512{\times}512$. Next, we separate and assign the speaker and listener audio using a visual-grounded speech separation model~\citep{iianet}. All videos are converted to 25~fps, and audio is resampled to 16~kHz. Additionally, we randomly select 50 videos from the talking-head dataset HDTF~\citep{hdtf} to evaluate the performance of talking-head generation.

\subsection{Implementation Details} 
We use the Adam optimizer~\citep{adam} with a learning rate of $10^{-4}$ and a batch size of 8. We retrain the motion latent auto-encoder from~\citet{float} on our dataset. The latent dimension is set to $d=512$. For our model $v_\theta$, we use 8 attention heads with a hidden dimension of $h$~$=$~$1024$ and 1D RoPE~\citep{rope}. It is trained with $N=50$ frames over $B=5$ blocks (i.e., 10 frames per block) and $l=2$ look-ahead frames. For audio encoding, we extract 12 multi-scale features from Wav2Vec2.0~\citep{wav2vec2}.
For motion sampling, we use 10 NFEs with the Euler solver and a classifier-free guidance~\citep{classifier_free}. All the experiments are conducted on a single NVIDIA H100 GPU.
% We will release our code and models upon publication.

%%%%%%%%%%%%%%%%%%%%%%%%%%%%%%%%%%%%%%%%%%%%%%%%%%%%%%%%%%%%%%%%%%%%%%%%%%%%%%%%%%%%%%%%%%%%%%%%
\begin{figure*}[t]
\vspace{-0.15in}
\begin{minipage}{1\textwidth}
    \centering
    \includegraphics[width=0.98\textwidth]{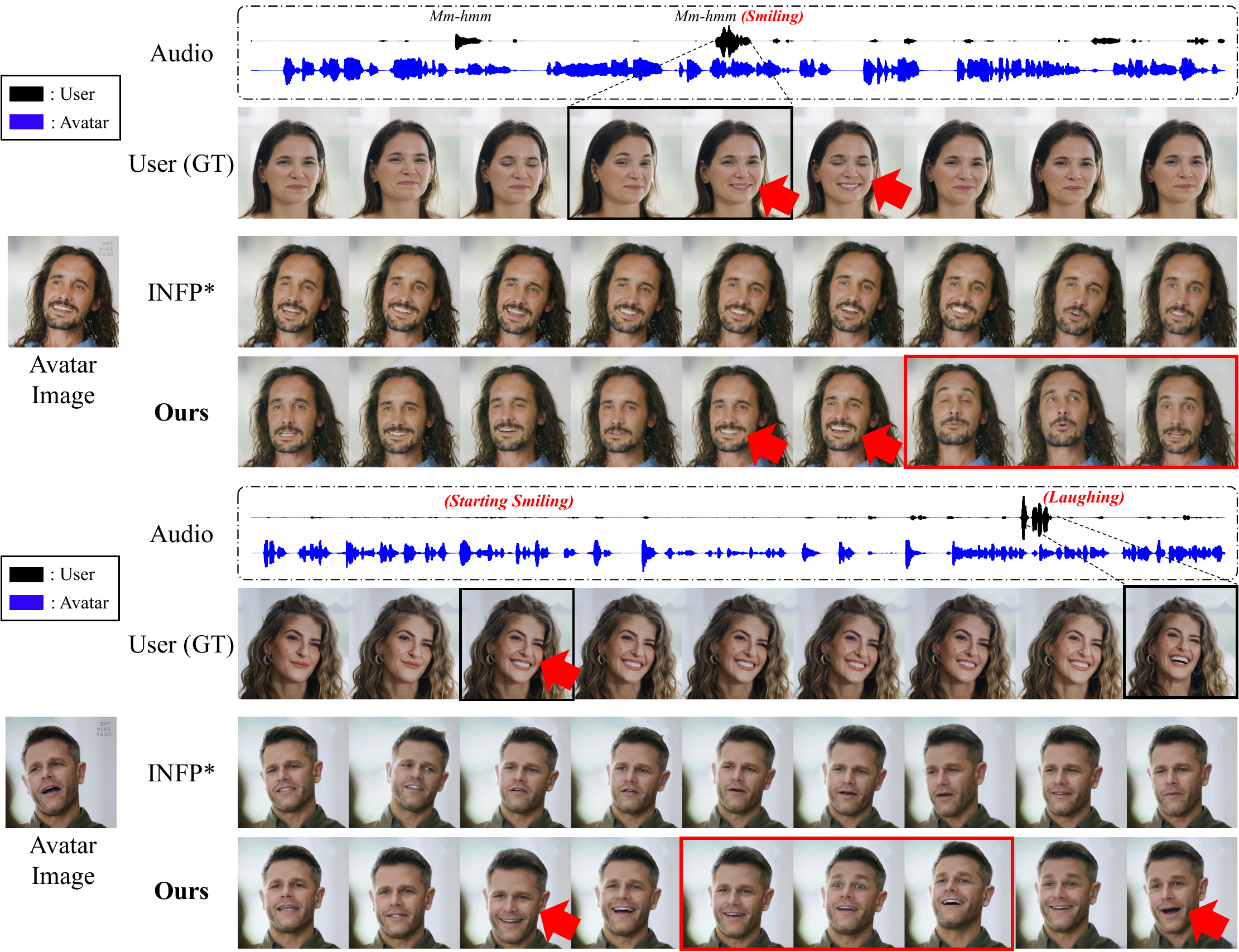}
    \vspace*{-2mm}
    \caption{\textbf{Qualitative comparison of interactive head avatar generation models} on the RealTalk~\citep{realtalk} dataset. Our model generates more reactive (red arrow) and expressive (red box) avatar motion compared to INFP*.
    We provide the videos in supplementary materials.
    }
    \label{fig:sota_compare}
\end{minipage}
\vspace*{-3mm}
\end{figure*}
%%%%%%%%%%%%%%%%%%%%%%%%%%%%%%%%%%%%%%%%%%%%%%%%%%%%%%%%%%%%%%%%%%%%%%%%%%%%%%%%%%%%%%%%%%%%%%%%

\subsection{Interactive Avatar Evaluation}
\paragraph{Metrics} We evaluate our model across five aspects of interactive avatar generation: Latency, Reactiveness, Motion Richness, Visual Quality, and Lip Synchronization.
For \textbf{Latency}, we assume that the pre-extracted audio features and measure motion generation time using 10 NFEs. For Reactiveness, we measure the motion synchronization between the user and the avatar by utilizing the residual Pearson correlation coefficients (rPCC) on facial expression (\textbf{rPCC-exp}) and head pose (\textbf{rPCC-pose})~\citep{emoca}. For Motion Richness, we measure the Similarity Index for diversity (\textbf{SID})~\citep{l2l} and variance (\textbf{Var}), following the previous studies~\citep{dim, infp}. For Visual Quality, we compute the Frechet inception distance (\textbf{FID})~\citep{fid} and the Frechet video distance (\textbf{FVD})~\citep{fvd} for the generated videos. We also assess the identity preservation using the cosine similarity of identity embeddings (\textbf{CSIM})~\citep{arcface} between the reference avatar image and the generated videos. In Lip Synchronization, we compute the lip sync error distance and confidence (\textbf{LSE-C} and \textbf{LSE-D}) using the generated video and the avatar's audio. Please refer to Appendix~\ref{sec:apx_exp_detail} for details on the metrics.

\paragraph{Comparison with Interactive Head Avatar}
We compare our model with the reproduced state-of-the-art dyadic talking avatar model, INFP*~\citep{infp}, since its official implementation is not publicly available. For reference, we also include a talking avatar generation model~\citep{float} which does not use user motion or audio.

In \cref{tab:quantitative_realtalk}, we provide a quantitative comparison on the RealTalk~\citep{realtalk} dataset. Avatar Forcing significantly outperforms INFP* in terms of Reactiveness and Motion Richness, indicating that our generated avatar is much more reactive and expressive compared to the baselines.
Notably, Avatar Forcing achieves a latency of 0.5s, enabling real-time interaction, and maintains Visual Quality and Lip Synchronization comparable to INFP*. In contrast, INFP*'s 3.4s latency makes it unsuitable for real-time applications.

We visualize the generated samples of ours and INFP* in \cref{fig:sota_compare}, where ours demonstrate more reactive (red arrow) and expressive motions (red box), compared to INFP*. 

\vspace{-0.05in}
\paragraph{Human Preference Study}
We conduct a human preference study comparing our model with the interactive head avatar generation model INFP*. We recruited 42 participants to evaluate 12 video sets using five perceptual metrics: \textbf{Reactiveness} measures how well the avatar reacts to user motion and audio, \textbf{Motion Richness} evaluates expressiveness of the avatar motion, \textbf{Verbal Alignment} evaluates the lip synchronization with the audio, \textbf{Non-verbal Alignment} assesses the non-verbal behaviors such as eye contact or nodding, and \textbf{Overall Preference}. We provide further details of the human study in the supplementary materials.

As shown in \cref{fig:human_eval}, our model is strongly preferred across all metrics, achieving over 80\% preference in overall quality. In particular,  our generated avatars exhibit expressive motion and strong non-verbal alignment, showing the effectiveness of the preference optimization (\cref{sec:dpo}).

\subsection{Comprehensive Analysis}
\paragraph{Comparison with Talking Head Avatar}
We evaluate the talking capability of the avatar. 
We compare Avatar Forcing with four state-of-the-art talking head avatar generation models:
SadTalker~\citep{sadtalker}, Hallo3~\citep{hallo3}, FLOAT~\citep{float}, and INFP$^*$. We measure the visual quality with \textbf{FID}, \textbf{FVD}, and \textbf{CSIM} metrics, and assess lip synchronization using \textbf{LSE-D} and \textbf{LSE-C} used in~\cref{tab:quantitative_realtalk}. We provide the quantitative results on the HDTF dataset~\citep{hdtf} in \cref{tab:compare_talking_hdtf}.
Avatar Forcing shows competitive performance on all metrics and achieves the best image and video quality. 
We provide visual examples in Appendix~\ref{sec:apx_exp_detail}.

\vspace{-0.075in}
\paragraph{Comparison with Listening Head Avatar}
We further evaluate the listening capability of our model. We compare Avatar Forcing with the listening head avatar generation models, including RLHG~\citep{rlhg}, L2L~\citep{l2l}, DIM~\citep{dim}, and INFP*, using the ViCo~\citep{rlhg}. 
We measure Frechet distance (FD) for the expression and pose, and rPCC, SID, and Var metrics used in~\cref{tab:quantitative_realtalk} but with respect to expression and head pose, respectively.
Since the baseline models are not publicly available, we take the results from DIM~\citep{dim}.
In \cref{tab:compare_listening_vico}, our model outperforms the baselines on almost all of the metrics. In particular, Avatar Forcing achieves the best user–avatar synchronized motion generation (rPCC).

%%%%%%%%%%%%%%%%%%%%%%%%%%%%%%%%%%%%%%%%%%%%%%%%%%%%%%%%%%%%%%%%%%%%%%%%%%%%%%%%%%%%%%%%%%%%%%%%
\begin{figure}[t]
\vspace{-0.1in}
\captionof{table}{
\textbf{Ablation study} on user motion and preference optimization. 
``w/ $\mathbf{m}_u$'' indicates whether the user motion latent $\mathbf{m}_u$ is provided as input to the model during both training and inference.
} \label{tab:ablation_realtalk}
\vspace{-0.0in}
\centering
\begin{minipage}{1\linewidth}
    \vspace*{-2mm}
    \resizebox{1.0\textwidth}{!}{
        \renewcommand{\arraystretch}{1.0}
        \renewcommand{\tabcolsep}{6pt}
        \begin{tabular}{c c c c c c}
        \toprule
         \multicolumn{2}{c}{Method} & \multicolumn{2}{c}{Reactiveness} & \multicolumn{2}{c}{Motion Richness} \\
         \cmidrule(l{2pt}r{2pt}){1-2}
        \cmidrule(l{2pt}r{2pt}){3-4}
        \cmidrule(l{2pt}r{2pt}){5-6}
        w/ $\mathbf{m}_u$ & DPO & rPCC-Exp $\downarrow$ & rPCC-Pose $\downarrow$ & SID $\uparrow$ & Var $\uparrow$\\
        \midrule        
        \redx & \redx       & 0.052 & 0.175 & 2.165 & 1.586\\
        \greencheck & \redx               & 0.042 & 0.146 & 2.236 & 1.408\\
        \midrule        
        \greencheck & \greencheck                & \textbf{0.003} & \textbf{0.036} & \textbf{2.442} & \textbf{1.734}\\
        \bottomrule
        \end{tabular}
        }
\end{minipage}
\vspace{-0.15in}
\end{figure}
%%%%%%%%%%%%%%%%%%%%%%%%%%%%%%%%%%%%%%%%%%%%%%%%%%%%%%%%%%%%%%%%%%%%%%%%%%%%%%%%%%%%%%%%%%%%%%%%

\subsection{Ablation Study} \label{sec:experiments:ablation}
In this section, we conduct ablation studies on (i) the necessity of using user motion in the Dual Motion Encoder and (ii) the importance of preference optimization. We provide further ablation studies, including the diffusion forcing and blockwise look-ahead attention mask in Appendix~\ref{sec:apx_exp_detail}.

\vspace{-0.05in}
\paragraph{User Motion} 
To validate the necessity of using user motion for interactive avatar generation, we compare our model with a variant that does not take user motion as input. As shown in \cref{tab:ablation_realtalk}, removing user motion leads to significantly less reactive behavior (Reactiveness) and reduced expressiveness in motion (Motion Richness).

\begin{figure}
\vspace{-0.1in}
\centering
\includegraphics[width=0.99\linewidth]{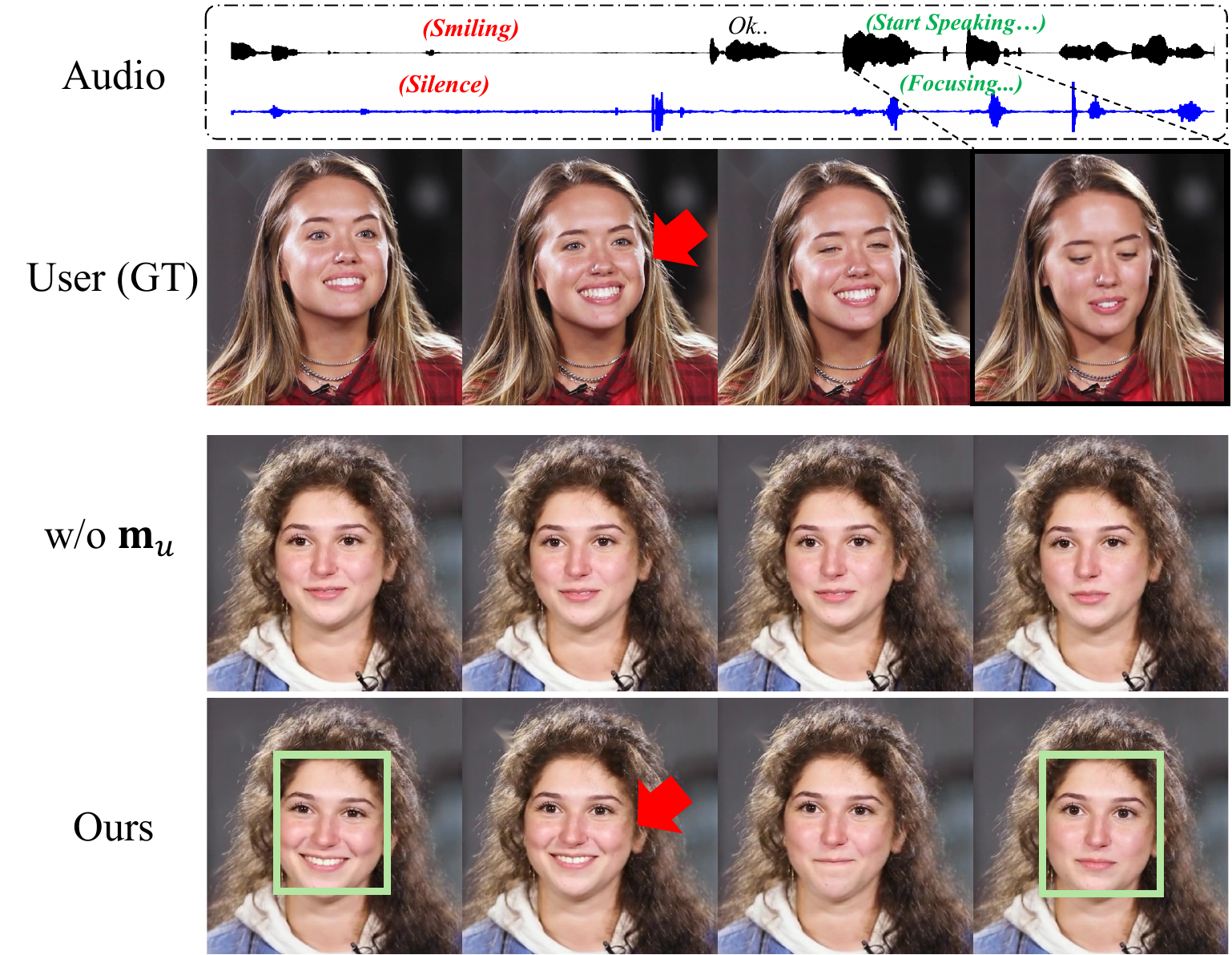}
\vspace*{-2mm}
\caption{\textbf{Ablation study on the user motion}. Without $\mathbf{m}_u$, the avatar remains static even when the user smiles. With $\mathbf{m}_u$, our model reacts by smiling after the user (red arrow) and shifting to a focused expression when the user begins speaking (green box).}
\label{fig:ablation_user}
\vspace*{-1mm}
\end{figure}

Furthermore, as visualized in~\cref{fig:ablation_user}, the model without user motion produces static behavior whenever the user's audio is silent. 
This occurs even in the presence of strong non-verbal cues, such as smiling, as the model cannot perceive visual signals. In contrast, our model, which uses user motion, generates a reactive avatar that naturally smiles right after the user smiles (\cref{fig:ablation_user} red arrow) and becomes more focused when the user speaks (\cref{fig:ablation_user} green box).

\begin{figure}[t]
    % \vspace*{-0.1in}
    \centering
    \includegraphics[width=0.99\linewidth]{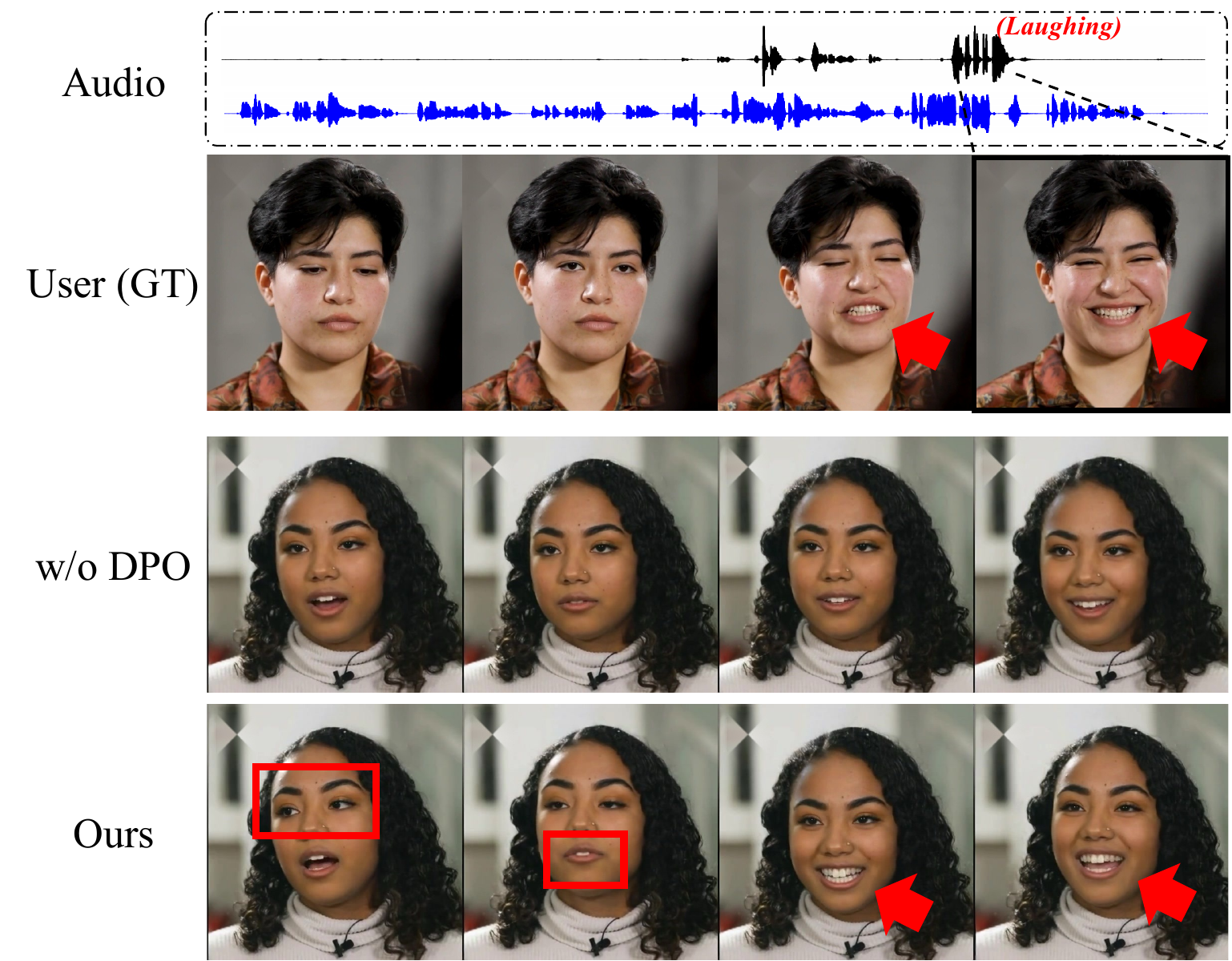}
    \vspace*{-2mm}
    \caption{\textbf{Ablation study on preference optimization}. Model fine-tuned with DPO produces more expressive motion (red box) and  reactive (red arrow)  compared to model without DPO.
    }\label{fig:ablation_dpo}   
    \vspace*{-0.18in}
\end{figure}

\vspace{-0.05in}
\paragraph{Direct Preference Optimization} 
To assess the impact of the preference optimization, we compare our model against a variant without fine-tuning. As shown in~\cref{tab:ablation_realtalk}, preference optimization significantly improves the Reactiveness metrics (rPCC-Exp and rPCC-Pose), which quantify user-avatar motion synchronization. It also substantially boosts the Motion Richness metrics (SID and Var), indicating more expressive and varied motion. As visualized in \cref{fig:ablation_dpo}, the model without preference optimization generates noticeably reduced diversity in facial expressions and head movement. It also exhibits weaker interaction, failing to respond to the user's smile. 

In contrast, our fine-tuned model generates an expressive avatar that shows natural head movement (\cref{fig:ablation_dpo} red box) and smiles more broadly along with the user (\cref{fig:ablation_dpo} red arrow). We provide further analysis of our DPO method in Appendix~\ref{sec:apx_exp_detail}.

%%%%%%%%%%%%%%%%%%%%%%%%%%%%%%%%%%%%%%%%%%%%%%%%%%%%%%%%%%%%%%%%%%%%%%%%%%%%%%%%%%%%%%%%%%%%%%%%
%%%%%%%%%%%%%%%%%%%%%%%%%%%%%%%%%%%%%%%%%%%%%%%%%%%%%%%%%%%%%%%%%%%%%%%%%%%%%%%%%%%%%%%%%%%%%%%%

%%%%%%%%%%%%%%%%%%%%%%%%%%%%%%%%%%%%%%%%%%%%%%%%%%%%%%%%%%%%%%%%%%%%%%%%%%%%%%%%%%%%%%%%%%%%%%%%
%%%%%%%%%%%%%%%%%%%%%%%%%%%%%%%%%%%%%%%%%%%%%%%%%%%%%%%%%%%%%%%%%%%%%%%%%%%%%%%%%%%%%%%%%%%%%%%%
\section{Conclusion} \label{sec:conclusion}
We proposed Avatar Forcing, a real-time interactive head avatar model based on diffusion forcing, that generates reactive and expressive motion using both the verbal and non-verbal user signals.Avatar Forcing takes a step toward truly interactive virtual avatars and opens new possibilities for real-time human-AI communication.

\noindent \textbf{Discussion}~We leave further discussion, including ethical considerations, limitations, and future work, in the Appendix~\ref{sec:apx_discussion}.

% \paragraph{Acknowledgment}
% This work is supported by Artificial intelligence industrial convergence cluster development project funded by the Ministry of Science and ICT(MSIT, Korea) \& Gwangju Metropolitan City.
\section{Acknowledgment}
This work was supported by Institute for Information \& communications Technology Planning \& Evaluation (IITP) grant funded by the Korea government (MSIT) (RS-2019-II190075, Artificial Intelligence Graduate School Program (KAIST)), National Research Foundation of Korea (NRF) grant funded by the Korea government (MSIT) (No. RS-2023-00256259), Center for Applied Research in Artificial Intelligence (CARAI) grant funded by DAPA and ADD (UD190031RD), and the InnoCORE program of the Ministry of Science and ICT (No. N10250156).
{
    \small
    \bibliographystyle{ieeenat_fullname}
    \bibliography{main}

@String(CVPR= {IEEE Conf. Comput. Vis. Pattern Recog.})

@String(AAAI = {AAAI})

@String(CVPR  = {CVPR})

@misc{synthesia,
  title={Synthesia},
  author={Synthesia},
  year={2025},
  howpublished={\url{https://www.synthesia.io/}},
}

@misc{hedra,
  title={Hedra Realtime Avatar},
  author={Hedra},
  year={2025},
  howpublished={\url{https://hedra.com/app/avatar}},
}

@misc{pika,
  title={Pika Audio-Driven Performance Model},
  author={Pika},
  year={2025},
  howpublished={\url{https://pika.art/api}},
}

@misc{pyscenedetect,
  title={PySceneDetect},
  author={PySceneDetect},
  year={2025},
  howpublished={\url{https://github.com/Breakthrough/PySceneDetect}},
}

@article{streamdit,
  title={Streamdit: Real-time streaming text-to-video generation},
  author={Kodaira, Akio and Hou, Tingbo and Hou, Ji and Tomizuka, Masayoshi and Zhao, Yue},
  journal={arXiv preprint arXiv:2507.03745},
  year={2025}
}

@inproceedings{iianet,
  title={IIANet: An Intra-and Inter-Modality Attention Network for Audio-Visual Speech Separation},
  author={Li, Kai and Yang, Runxuan and Sun, Fuchun and Hu, Xiaolin},
  booktitle={International Conference on Machine Learning},
  year={2024},
}

@article{ouyang2022training,
  title={Training language models to follow instructions with human feedback},
  author={Ouyang, Long and Wu, Jeffrey and Jiang, Xu and Almeida, Diogo and Wainwright, Carroll and Mishkin, Pamela and Zhang, Chong and Agarwal, Sandhini and Slama, Katarina and Ray, Alex and others},
  journal={Advances in Neural Information Processing Systems},
  year={2022}
}

@inproceedings{wallace2024diffusiondpo,
  title={Diffusion model alignment using direct preference optimization},
  author={Wallace, Bram and Dang, Meihua and Rafailov, Rafael and Zhou, Linqi and Lou, Aaron and Purushwalkam, Senthil and Ermon, Stefano and Xiong, Caiming and Joty, Shafiq and Naik, Nikhil},
  booktitle={Conference on Computer Vision and Pattern Recognition},
  year={2024}
}

@article{rafailov2023dpo,
  title={Direct preference optimization: Your language model is secretly a reward model},
  author={Rafailov, Rafael and Sharma, Archit and Mitchell, Eric and Manning, Christopher D and Ermon, Stefano and Finn, Chelsea},
  journal={Advances in Neural Information Processing Systems},
  year={2023}
}

@inproceedings{diffusion_listening,
    author    = {Wang, Yinuo and Fan, Yanbo and Wang, Xuan and Yu, Guo and Wang, Fei},
    title     = {Diffusion-based Realistic Listening Head Generation via Hybrid Motion Modeling},
  booktitle={Conference on Computer Vision and Pattern Recognition},
    year      = {2025},
}

@inproceedings{emoca,
  title={Emoca: Emotion driven monocular face capture and animation},
  author={Dan{\v{e}}{\v{c}}ek, Radek and Black, Michael J and Bolkart, Timo},
  booktitle={Conference on Computer Vision and Pattern Recognition},
  year={2022}
}

@inproceedings{l2l,
  title={Learning to listen: Modeling non-deterministic dyadic facial motion},
  author={Ng, Evonne and Joo, Hanbyul and Hu, Liwen and Li, Hao and Darrell, Trevor and Kanazawa, Angjoo and Ginosar, Shiry},
  booktitle={Conference on Computer Vision and Pattern Recognition},
  year={2022}
}

@article{vqvae,
  title={Neural discrete representation learning},
  author={Van Den Oord, Aaron and Vinyals, Oriol and others},
  journal={Advances in Neural Information Processing Systems},
  year={2017}
}

@article{spectre,
  title={Visual speech-aware perceptual 3d facial expression reconstruction from videos},
  author={Filntisis, Panagiotis P and Retsinas, George and Paraperas-Papantoniou, Foivos and Katsamanis, Athanasios and Roussos, Anastasios and Maragos, Petros},
  journal={arXiv preprint arXiv:2207.11094},
  year={2022}
}

@article{genie3,
  title         = {Genie 3: A New Frontier for World Models},
  author        = {Philip J. Ball and Jakob Bauer and Frank Belletti and Bethanie Brownfield and Ariel Ephrat and Shlomi Fruchter and Agrim Gupta and Kristian Holsheimer and Aleksander Holynski and Jiri Hron and Christos Kaplanis and Marjorie Limont and Matt McGill and Yanko Oliveira and Jack Parker-Holder and Frank Perbet and Guy Scully and Jeremy Shar and Stephen Spencer and Omer Tov and Ruben Villegas and Emma Wang and Jessica Yung and Cip Baetu and Jordi Berbel and David Bridson and Jake Bruce and Gavin Buttimore and Sarah Chakera and Bilva Chandra and Paul Collins and Alex Cullum and Bogdan Damoc and Vibha Dasagi and Maxime Gazeau and Charles Gbadamosi and Woohyun Han and Ed Hirst and Ashyana Kachra and Lucie Kerley and Kristian Kjems and Eva Knoepfel and Vika Koriakin and Jessica Lo and Cong Lu and Zeb Mehring and Alex Moufarek and Henna Nandwani and Valeria Oliveira and Fabio Pardo and Jane Park and Andrew Pierson and Ben Poole and Helen Ran and Tim Salimans and Manuel Sanchez and Igor Saprykin and Amy Shen and Sailesh Sidhwani and Duncan Smith and Joe Stanton and Hamish Tomlinson and Dimple Vijaykumar and Luyu Wang and Piers Wingfield and Nat Wong and Keyang Xu and Christopher Yew and Nick Young and Vadim Zubov and Douglas Eck and Dumitru Erhan and Koray Kavukcuoglu and Demis Hassabis and Zoubin Gharamani and Raia Hadsell and A{\"a}ron van den Oord and Inbar Mosseri and Adrian Bolton and Satinder Singh and Tim Rockt{\"a}schel},
  year          = {2025},
  url           = {}
}

@article{magicinfinite,
  title={Magicinfinite: Generating infinite talking videos with your words and voice},
  author={Yi, Hongwei and Ye, Tian and Shao, Shitong and Yang, Xuancheng and Zhao, Jiantong and Guo, Hanzhong and Wang, Terrance and Yin, Qingyu and Xie, Zeke and Zhu, Lei and others},
  journal={arXiv preprint arXiv:2503.05978},
  year={2025}
}

@article{mocha,
  title={Mocha: Towards movie-grade talking character synthesis},
  author={Wei, Cong and Sun, Bo and Ma, Haoyu and Hou, Ji and Juefei-Xu, Felix and He, Zecheng and Dai, Xiaoliang and Zhang, Luxin and Li, Kunpeng and Hou, Tingbo and others},
  journal={arXiv preprint arXiv:2503.23307},
  year={2025}
}

@inproceedings{hunyuanportrait,
  title={Hunyuanportrait: Implicit condition control for enhanced portrait animation},
  author={Xu, Zunnan and Yu, Zhentao and Zhou, Zixiang and Zhou, Jun and Jin, Xiaoyu and Hong, Fa-Ting and Ji, Xiaozhong and Zhu, Junwei and Cai, Chengfei and Tang, Shiyu and others},
  booktitle={Conference on Computer Vision and Pattern Recognition},
  year={2025}
}

@article{talkingmachines,
  title={TalkingMachines: Real-Time Audio-Driven FaceTime-Style Video via Autoregressive Diffusion Models},
  author={Low, Chetwin and Wang, Weimin},
  journal={arXiv preprint arXiv:2506.03099},
  year={2025}
}

@article{omnihuman,
  title={Omnihuman-1: Rethinking the scaling-up of one-stage conditioned human animation models},
  author={Lin, Gaojie and Jiang, Jianwen and Yang, Jiaqi and Zheng, Zerong and Liang, Chao},
  journal={arXiv preprint arXiv:2502.01061},
  year={2025}
}

@inproceedings{mfr_net,
  title={Mfr-net: Multi-faceted responsive listening head generation via denoising diffusion model},
  author={Liu, Jin and Wang, Xi and Fu, Xiaomeng and Chai, Yesheng and Yu, Cai and Dai, Jiao and Han, Jizhong},
  booktitle={Association for Computing Machinery International Conference on Multimedia},
  year={2023}
}

@article{ditailistener,
  title={DiTaiListener: Controllable High Fidelity Listener Video Generation with Diffusion},
  author={Siniukov, Maksim and Chang, Di and Tran, Minh and Gong, Hongkun and Chaubey, Ashutosh and Soleymani, Mohammad},
  journal={arXiv preprint arXiv:2504.04010},
  year={2025}
}

@inproceedings{pch,
  title={Perceptual conversational head generation with regularized driver and enhanced renderer},
  author={Huang, Ailin and Huang, Zhewei and Zhou, Shuchang},
  booktitle={Association for Computing Machinery International Conference on Multimedia},
  year={2022}
}

@article{realtalk,
  title={Affective faces for goal-driven dyadic communication},
  author={Geng, Scott and Teotia, Revant and Tendulkar, Purva and Menon, Sachit and Vondrick, Carl},
  journal={arXiv preprint arXiv:2301.10939},
  year={2023}
}

@inproceedings{customlistener,
  title={Customlistener: Text-guided responsive interaction for user-friendly listening head generation},
  author={Liu, Xi and Guo, Ying and Zhen, Cheng and Li, Tong and Ao, Yingying and Yan, Pengfei},
  booktitle={Conference on Computer Vision and Pattern Recognition},
  year={2024}
}

@inproceedings{rlhg,
  title={Responsive listening head generation: a benchmark dataset and baseline},
  author={Zhou, Mohan and Bai, Yalong and Zhang, Wei and Yao, Ting and Zhao, Tiejun and Mei, Tao},
  booktitle={European Conference on Computer Vision},
  year={2022},
}

@inproceedings{hallo3,
  title={Hallo3: Highly dynamic and realistic portrait image animation with video diffusion transformer},
  author={Cui, Jiahao and Li, Hui and Zhan, Yun and Shang, Hanlin and Cheng, Kaihui and Ma, Yuqi and Mu, Shan and Zhou, Hang and Wang, Jingdong and Zhu, Siyu},
  booktitle={Conference on Computer Vision and Pattern Recognition},
  year={2025}
}

@inproceedings{diffusion_forcing,
  title={Diffusion forcing: Next-token prediction meets full-sequence diffusion},
  author={Chen, Boyuan and Mart{\'\i} Mons{\'o}, Diego and Du, Yilun and Simchowitz, Max and Tedrake, Russ and Sitzmann, Vincent},
  booktitle={Advances in Neural Information Processing Systems},
  year={2024}
}

@inproceedings{causvid,
  title={From slow bidirectional to fast autoregressive video diffusion models},
  author={Yin, Tianwei and Zhang, Qiang and Zhang, Richard and Freeman, William T and Durand, Fredo and Shechtman, Eli and Huang, Xun},
  booktitle={Conference on Computer Vision and Pattern Recognition},
  year={2025}
}

@inproceedings{self_forcing,
  title={Self Forcing: Bridging the Train-Test Gap in Autoregressive Video Diffusion},
  author={Huang, Xun and Li, Zhengqi and He, Guande and Zhou, Mingyuan and Shechtman, Eli},
  booktitle={Advances in Neural Information Processing Systems},
  year={2025}
}

@inproceedings{float,
  title={Float: Generative motion latent flow matching for audio-driven talking portrait},
  author={Ki, Taekyung and Min, Dongchan and Chae, Gyeongsu},
  booktitle={International Conference on Computer Vision},
  year={2025}
}

@inproceedings{infp,
  title={INFP: Audio-driven interactive head generation in dyadic conversations},
  author={Zhu, Yongming and Zhang, Longhao and Rong, Zhengkun and Hu, Tianshu and Liang, Shuang and Ge, Zhipeng},
  booktitle={Conference on Computer Vision and Pattern Recognition},
  year={2025}
}

@inproceedings{arig,
    author    = {Guo, Ying and Liu, Xi and Zhen, Cheng and Yan, Pengfei and Wei, Xiaoming},
    title     = {ARIG: Autoregressive Interactive Head Generation for Real-time Conversations},
    booktitle = {International Conference on Computer Vision},
    year      = {2025},
    
}

@inproceedings{dim,
  title={Dim: Dyadic interaction modeling for social behavior generation},
  author={Tran, Minh and Chang, Di and Siniukov, Maksim and Soleymani, Mohammad},
  booktitle={European Conference on Computer Vision},
  year={2024},
}

@inproceedings{historydiffusion,
  title={History-Guided Video Diffusion}, 
  author={Kiwhan Song and Boyuan Chen and Max Simchowitz and Yilun Du and Russ Tedrake and Vincent Sitzmann},
  year={2025},
  booktitle    = {International Conference on Machine Learning},
}

@inproceedings{ldm,
  title={High-resolution image synthesis with latent diffusion models},
  author={Rombach, Robin and Blattmann, Andreas and Lorenz, Dominik and Esser, Patrick and Ommer, Bj{\"o}rn},
  booktitle={Conference on Computer Vision and Pattern Recognition},
  year={2022}
}

@article{classifier_free,
  title={Classifier-free diffusion guidance},
  author={Ho, Jonathan and Salimans, Tim},
  journal={arXiv preprint arXiv:2207.12598},
  year={2022}
}

@inproceedings{sde,
  title={Score-based generative modeling through stochastic differential equations},
  author={Song, Yang and Sohl-Dickstein, Jascha and Kingma, Diederik P and Kumar, Abhishek and Ermon, Stefano and Poole, Ben},
  booktitle={International Conference on Learning Representations},
  year={2021}
}

@inproceedings{cfm,
  title={Flow matching for generative modeling},
  author={Lipman, Yaron and Chen, Ricky TQ and Ben-Hamu, Heli and Nickel, Maximilian and Le, Matt},
  booktitle={International Conference on Learning Representations},
  year={2023}
}

@article{wav2vec2,
  title={wav2vec 2.0: A framework for self-supervised learning of speech representations},
  author={Baevski, Alexei and Zhou, Yuhao and Mohamed, Abdelrahman and Auli, Michael},
  journal={Advances in Neural Information Processing Systems},
  volume={33},
  pages={12449--12460},
  year={2020}
}

@inproceedings{wav2lip,
  title={A lip sync expert is all you need for speech to lip generation in the wild},
  author={Prajwal, KR and Mukhopadhyay, Rudrabha and Namboodiri, Vinay P and Jawahar, CV},
  booktitle={Association for Computing Machinery International Conference on Multimedia},
  year={2020}
}

@inproceedings{syncnet,
  title={Out of time: automated lip sync in the wild},
  author={Chung, Joon Son and Zisserman, Andrew},
  booktitle={Asian Conference on Computer Vision},
  year={2016}
}

@inproceedings{anitalker,
  title={Anitalker: animate vivid and diverse talking faces through identity-decoupled facial motion encoding},
  author={Liu, Tao and Chen, Feilong and Fan, Shuai and Du, Chenpeng and Chen, Qi and Chen, Xie and Yu, Kai},
  booktitle={Proceedings of the 32nd ACM International Conference on Multimedia},
  pages={6696--6705},
  year={2024}
}

@article{lia,
  title={Latent image animator: Learning to animate images via latent space navigation},
  author={Wang, Yaohui and Yang, Di and Bremond, Francois and Dantcheva, Antitza},
  journal={arXiv preprint arXiv:2203.09043},
  year={2022}
}

@inproceedings{sadtalker,
  title={Sadtalker: Learning realistic 3d motion coefficients for stylized audio-driven single image talking face animation},
  author={Zhang, Wenxuan and Cun, Xiaodong and Wang, Xuan and Zhang, Yong and Shen, Xi and Guo, Yu and Shan, Ying and Wang, Fei},
  booktitle={Conference on Computer Vision and Pattern Recognition},
  year={2023}
}

@inproceedings{stylelipsync,
  title={StyleLipSync: Style-based personalized lip-sync video generation},
  author={Ki, Taekyung and Min, Dongchan},
  booktitle={International Conference on Computer Vision},
  year={2023}
}

@article{liveportrait,
  title={LivePortrait: Efficient Portrait Animation with Stitching and Retargeting Control},
  author={Guo, Jianzhu and Zhang, Dingyun and Liu, Xiaoqiang and Zhong, Zhizhou and Zhang, Yuan and Wan, Pengfei and Zhang, Di},
  journal={arXiv preprint arXiv:2407.03168},
  year={2024}
}

@inproceedings{ar_wo_vq,
  title={Autoregressive image generation without vector quantization},
  author={Li, Tianhong and Tian, Yonglong and Li, He and Deng, Mingyang and He, Kaiming},
  booktitle={Advances in Neural Information Processing Systems},
  year={2024}
}

@article{rope,
  title={Roformer: Enhanced transformer with rotary position embedding},
  author={Su, Jianlin and Ahmed, Murtadha and Lu, Yu and Pan, Shengfeng and Bo, Wen and Liu, Yunfeng},
  journal={Neurocomputing},
  year={2024},
}

@inproceedings{dit,
  title={Scalable diffusion models with transformers},
  author={Peebles, William and Xie, Saining},
  booktitle={International Conference on Computer Vision},
  year={2023}
}

@article{pixart_a,
  title={Pixart-$\alpha$: Fast training of diffusion transformer for photorealistic text-to-image synthesis},
  author={Chen, Junsong and Yu, Jincheng and Ge, Chongjian and Yao, Lewei and Xie, Enze and Wu, Yue and Wang, Zhongdao and Kwok, James and Luo, Ping and Lu, Huchuan and others},
  journal={arXiv preprint arXiv:2310.00426},
  year={2023}
}

@article{vasa_1,
  title={Vasa-1: Lifelike audio-driven talking faces generated in real time},
  author={Xu, Sicheng and Chen, Guojun and Guo, Yu-Xiao and Yang, Jiaolong and Li, Chong and Zang, Zhenyu and Zhang, Yizhong and Tong, Xin and Guo, Baining},
  journal={Advances in Neural Information Processing Systems},
  year={2024}
}

@inproceedings{emo,
  title={Emo: Emote portrait alive generating expressive portrait videos with audio2video diffusion model under weak conditions},
  author={Tian, Linrui and Wang, Qi and Zhang, Bang and Bo, Liefeng},
  booktitle={European Conference on Computer Vision},
  pages={244--260},
  year={2024}
}

@article{loopy,
  title={Loopy: Taming Audio-Driven Portrait Avatar with Long-Term Motion Dependency},
  author={Jiang, Jianwen and Liang, Chao and Yang, Jiaqi and Lin, Gaojie and Zhong, Tianyun and Zheng, Yanbo},
  journal={arXiv preprint arXiv:2409.02634},
  year={2024}
}

@article{ddpm,
  title={Denoising diffusion probabilistic models},
  author={Ho, Jonathan and Jain, Ajay and Abbeel, Pieter},
  journal={Advances in neural information processing systems},
  volume={33},
  pages={6840--6851},
  year={2020}
}

@article{rectflow,
  title={Flow straight and fast: Learning to generate and transfer data with rectified flow},
  author={Liu, Xingchao and Gong, Chengyue and Liu, Qiang},
  journal={arXiv preprint arXiv:2209.03003},
  year={2022}
}

@inproceedings{instructpix2pix,
  title={Instructpix2pix: Learning to follow image editing instructions},
  author={Brooks, Tim and Holynski, Aleksander and Efros, Alexei A},
  booktitle={Proceedings of the IEEE/CVF Conference on Computer Vision and Pattern Recognition (CVPR)},
  pages={18392--18402},
  year={2023}
}

@inproceedings{echomimic,
  title={Echomimic: Lifelike audio-driven portrait animations through editable landmark conditions},
  author={Chen, Zhiyuan and Cao, Jiajiong and Chen, Zhiquan and Li, Yuming and Ma, Chenguang},
  booktitle={Proceedings of the AAAI Conference on Artificial Intelligence},
  volume={39},
  number={3},
  pages={2403--2410},
  year={2025}
}

@article{makeittalk,
  title={Makelttalk: speaker-aware talking-head animation},
  author={Zhou, Yang and Han, Xintong and Shechtman, Eli and Echevarria, Jose and Kalogerakis, Evangelos and Li, Dingzeyu},
  journal      = {Association for Computing Machinery Transactions on Graphics},

  year={2020},
}

@inproceedings{stylesync,
  title={Stylesync: High-fidelity generalized and personalized lip sync in style-based generator},
  author={Guan, Jiazhi and Zhang, Zhanwang and Zhou, Hang and Hu, Tianshu and Wang, Kaisiyuan and He, Dongliang and Feng, Haocheng and Liu, Jingtuo and Ding, Errui and Liu, Ziwei and others},
  booktitle={Conference on Computer Vision and Pattern Recognition},

  year={2023}
}

@misc{fid,
  author={Maximilian Seitzer},
  title={{pytorch-fid: FID Score for PyTorch}},
  month={August},
  year={2020},
  note={Version 0.3.0},
  url={\url{https://github.com/mseitzer/pytorch-fid}},
}

@article{fvd,
  title={Towards accurate generative models of video: A new metric \& challenges},
  author={Unterthiner, Thomas and Van Steenkiste, Sjoerd and Kurach, Karol and Marinier, Raphael and Michalski, Marcin and Gelly, Sylvain},
  journal={arXiv preprint arXiv:1812.01717},
  year={2018}
}

@inproceedings{arcface,
  title={Arcface: Additive angular margin loss for deep face recognition},
  author={Deng, Jiankang and Guo, Jia and Xue, Niannan and Zafeiriou, Stefanos},
  booktitle={Conference on Computer Vision and Pattern Recognition},
  year={2019}
}

@article{keysync,
  title={Keysync: A robust approach for leakage-free lip synchronization in high resolution},
  author={Bigata, Antoni and Mira, Rodrigo and Bounareli, Stella and Stypu{\l}kowski, Micha{\l} and Vougioukas, Konstantinos and Petridis, Stavros and Pantic, Maja},
  journal={arXiv preprint arXiv:2505.00497},
  year={2025}
}

@inproceedings{bfm,
  title={Accurate 3d face reconstruction with weakly-supervised learning: From single image to image set},
  author={Deng, Yu and Yang, Jiaolong and Xu, Sicheng and Chen, Dong and Jia, Yunde and Tong, Xin},
  booktitle={Conference on Computer Vision and Pattern Recognition Workshops},
  year={2019}
}

@article{adam,
  title={Adam: A method for stochastic optimization},
  author={Kingma, Diederik P},
  journal={arXiv preprint arXiv:1412.6980},
  year={2014}
}

@inproceedings{hdtf,
  title={Flow-guided one-shot talking face generation with a high-resolution audio-visual dataset},
  author={Zhang, Zhimeng and Li, Lincheng and Ding, Yu and Fan, Changjie},
  booktitle={Conference on Computer Vision and Pattern Recognition},

  year={2021}
}

@inproceedings{face_alignment,
  title={How far are we from solving the 2D \& 3D Face Alignment problem? (and a dataset of 230,000 3D facial landmarks)},
  author={Bulat, Adrian and Tzimiropoulos, Georgios},
  booktitle={International Conference on Computer Vision},
  year={2017}
}

@inproceedings{megaportrait,
  title={Megaportraits: One-shot megapixel neural head avatars},
  author={Drobyshev, Nikita and Chelishev, Jenya and Khakhulin, Taras and Ivakhnenko, Aleksei and Lempitsky, Victor and Zakharov, Egor},
  booktitle={Association for Computing Machinery International Conference on Multimedia},
  year={2022}
}

@inproceedings{emoportraits,
  title={EMOPortraits: Emotion-enhanced Multimodal One-shot Head Avatars},
  author={Drobyshev, Nikita and Casademunt, Antoni Bigata and Vougioukas, Konstantinos and Landgraf, Zoe and Petridis, Stavros and Pantic, Maja},
  booktitle={Conference on Computer Vision and Pattern Recognition},
  year={2024}
}
}
\maketitlesupplementary
\appendix 
\paragraph{Organization} The appendix is organized as follows: In Appendix~\ref{sec:apx_background}, we provide the additional backgrounds of our work. We describe the details of our model architecture in Appendix~\ref{sec:apx_architecture} and the preference optimization method in Appendix~\ref{sec:apx_dpo}. Experimental details are presented in Appendix~\ref{sec:apx_exp_detail}, and further discussion is provided in Appendix~\ref{sec:apx_discussion}.

\section{Background}\label{sec:apx_background}
\paragraph{Flow Matching} Flow matching~\citep{cfm, rectflow} is a generative model that transforms a simple prior distribution $p_0$, for example, a Gaussian distribution, into the target data distribution $p_1$ via an ordinary differential equation (ODE):
\begin{equation}
    \frac{d x_t}{dt} = v_{\theta}(x_t, t), ~~ t \in [0, 1], \label{eq:ode}
\end{equation}
where for fixed $x_0$ $\sim$ $p_0$ and $x_1$ $\sim$ $p_1$, the intermediate sample is a linear interpolation $x_t = tx_1 + (1-t) x_0$. The training objective of flow matching is to regress the vector field $v_{\theta}$ toward the target vector field $v_t = x_1 - x_0$:
\begin{equation}
    \loss_{FM}(\theta) = \mathbb{E}_{t, x_t} \left[\| v_{\theta}(x_t, t) - (x_1 - x_0)\|\right]. \label{eq:train_fm}
\end{equation}
It then generates target samples by solving Eq.~\eqref{eq:ode}. Note that flow matching can be interpreted as a diffusion model~\citep{ddpm, sde} where a noise schedule follows the linear trajectory between the prior and the target data.

\section{Details on Model Architecture}\label{sec:apx_architecture}
In Appendix~\ref{sec:apx_motion_ae}, we provide more details on the motion latent auto-encoder. In Appendix~\ref{sec:apx_vector_field_model}, we provide more details on the vector field predictor $v_\theta$.

\subsection{Motion Latent Auto-encoder}\label{sec:apx_motion_ae}

\begin{figure}
    \centering
    \includegraphics[width=\linewidth]{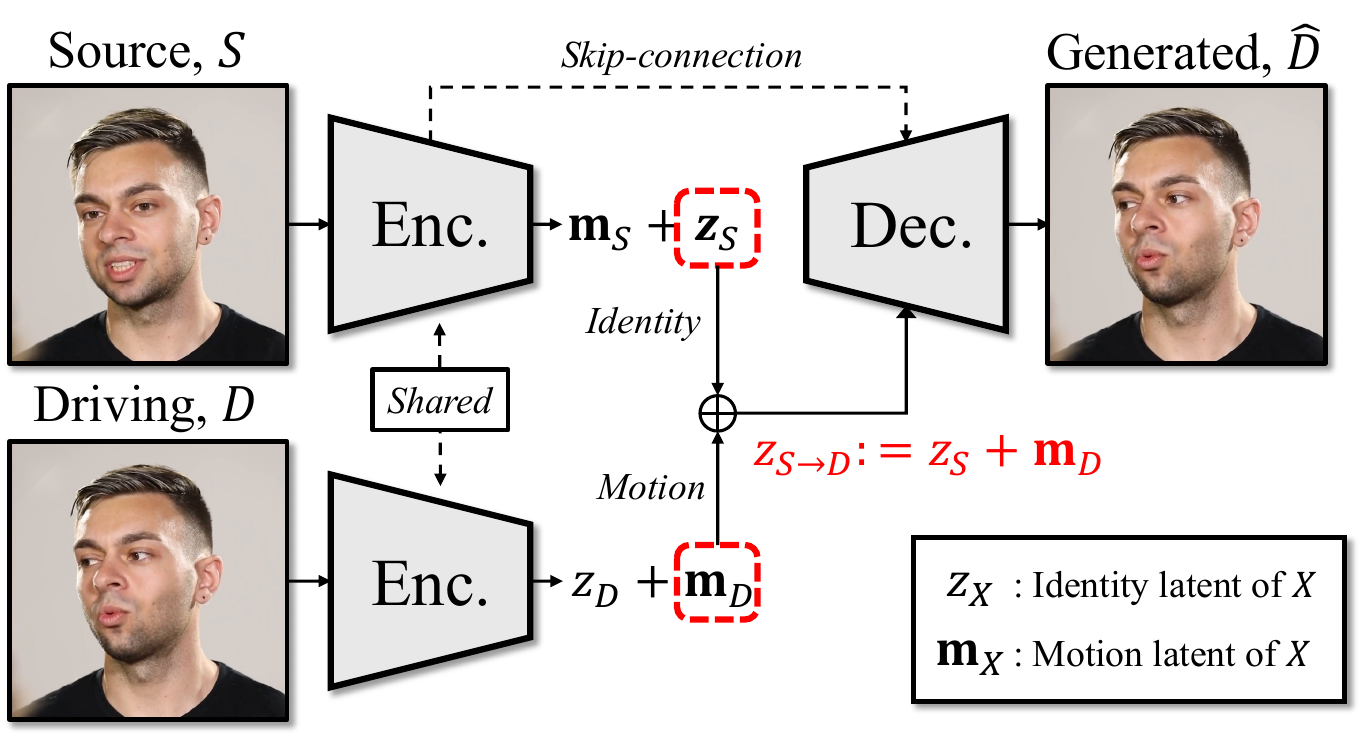}
    \vspace*{-5mm}
    \caption{\textbf{Overview of Motion Latent Encoder}. It encodes an image into a latent vector that has explicit identity-motion decomposition.\label{fig:motion_latent_ae}}
    \vspace*{-3mm}
\end{figure}

In~\cref{fig:motion_latent_ae}, we show an overview of the motion latent auto-encoder. It encodes an image into a latent vector that can be decomposed into an identity representation (i.e., appearance) and the motion representation. This auto-encoder is trained to reconstruct a driving image using a source image that shares the same identity. During each training iteration, the encoder encodes two images $S$ and $D$ drawn from the same video clip, and computes the $z_{S \to D} := z_S + \mathbf{m}_D$ that transforms the source image into the reconstructed $\hat{D}$. This explicit decomposition yields a compact motion representation, enabling fast motion generation.

We train this auto-encoder on our dataset, following the training objective described in the original paper. For more details, including the property of this latent space, training details, please refer to~\citep{lia, float}. 
%%%%%%%%%%%%%%%%%%%%%%%%%%%%%%%%%%%%%%%%%%%%%%%%%%%%%%%%%%%%%%%%%%%%%%%%%%%%%%%%%%%%%%%%%%%%%%%%
%%%%%%%%%%%%%%%%%%%%%%%%%%%%%%%%%%%%%%%%%%%%%%%%%%%%%%%%%%%%%%%%%%%%%%%%%%%%%%%%%%%%%%%%%%%%%%%%

%%%%%%%%%%%%%%%%%%%%%%%%%%%%%%%%%%%%%%%%%%%%%%%%%%%%%%%%%%%%%%%%%%%%%%%%%%%%%%%%%%%%%%%%%%%%%%%%
\begin{figure}
    \centering
    \includegraphics[width=\linewidth]{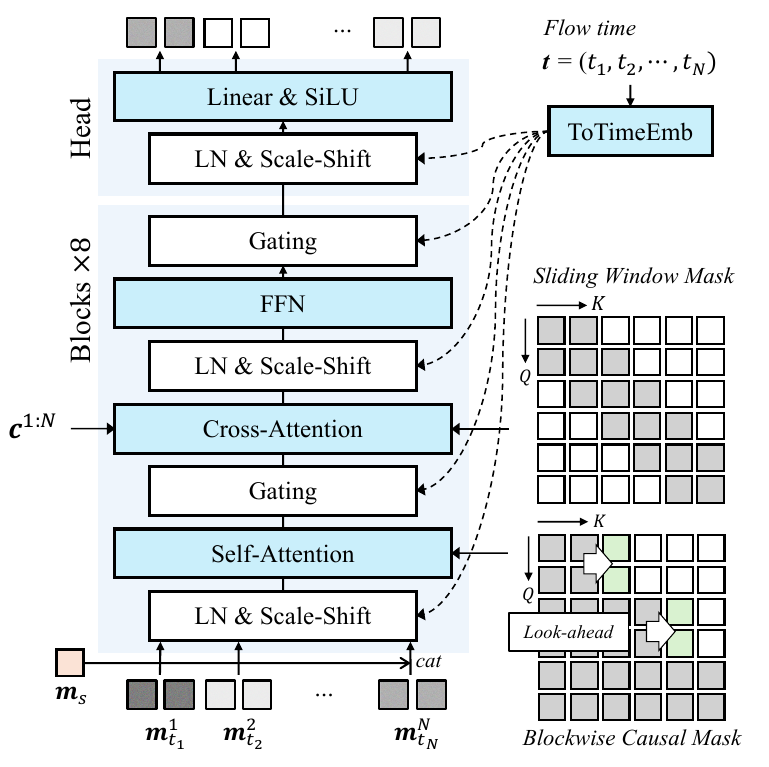}
    \vspace*{-7mm}
    \caption{\textbf{Detailed architecture for Motion Generator in $v_{\theta}$}.\label{fig:motion_generator}}
\end{figure}
%%%%%%%%%%%%%%%%%%%%%%%%%%%%%%%%%%%%%%%%%%%%%%%%%%%%%%%%%%%%%%%%%%%%%%%%%%%%%%%%%%%%%%%%%%%%%%%%

\subsection{Vector Field Model $v_\theta$}\label{sec:apx_vector_field_model}

\paragraph{Model Architecture}
The model $v_{\theta}$ comprises two main components: the Dual Motion Encoder and the Causal DFoT Motion Generator. The Dual Motion Encoder unifies three multimodal inputs through a cross-attention layer, computed as $\text{softmax}\left(\frac{QK^\top}{\sqrt{d}}\right)V$, where

\begin{equation}
    Q = \mathbf{q}W_q, \qquad
    K = \mathbf{k}W_k, \qquad
    V = \mathbf{v}W_v,
\end{equation}

and $W_q, W_k, W_v \in \mathbb{R}^{d \times d}$ are learnable projection matrices for the query $\mathbf{q}$, key $\mathbf{k}$, and value $\mathbf{v} \in \mathbb{R}^{N \times d}$, respectively ($N$ is the number of latents). In the first cross-attention layer, the encoder captures holistic verbal and non-verbal user motion by using the user motion latent $\mathbf{m}_u$ as the query. In the second layer, it integrates this aligned user motion with the avatar’s audio by taking the avatar audio as the query. We use four attention heads, each with a hidden dimension of $d = 512$, for both cross-attention layers.

In~\cref{fig:motion_generator}, we provide a detailed architecture for the motion generator. It consists of eight DFoT transformer blocks followed by a transformer head. Specifically, in each DFoT block, noisy latents $(\mathbf{m}_{t_1}^1, \mathbf{m}_{t_2}^2, \cdots, \mathbf{m}_{t_N}^N)$ are modulated by the flow time $\mathbf{t} = (t_1, t_2, \cdots, t_N)$ through a shared AdaLN scale-shift coefficients (ToTimeEmb layer)~\citep{pixart_a}. 

For the attention modules, we use Blockwise Causal Look-ahead Mask (Eq.~\eqref{eq:lookahead_mask}) in self-attention and a Sliding-window Attention Mask for aligning the driving signal $\mathbf{c}^{1:N}$ to the noisy latents. Specifically, we introduce the Blockwise Casual Look-ahead Mask to ensure the causal motion generation in our motion latent space, which significantly improves the temporal consistency of the generated video, as demonstrated in Appendix~\ref{sec:apx_additional_ablation}. Unlike the recent video diffusion models that employ a spatio-temporal compression module~\citep{causvid, self_forcing} where each latent correlates to multiple video frames by the compression rate (e.g., 4$\times$ or 8$\times$), our motion latent has one-to-one correspondence with each frame in pixel space. Under this setting, a simple (blockwise) causal mask alone produces the temporal inconsistencies across the frames or blocks. 

\section{Details on Preference Optimization}\label{sec:apx_dpo}

\paragraph{Training Objective Formulation}
Inspired by DiffusionDPO~\citep{wallace2024diffusiondpo}, we formulate the training objective $\loss_{DPO}$ in the context of diffusion forcing~\citep{diffusion_forcing}. Let $(\mathbf{m}^l, \mathbf{m}^w)$ denote a pair of less-preferred and preferred motion latents, each consisting of $N$ frames, where $\mathbf{m}^l \coloneq (\mathbf{m}^{l, n})_{n=1}^{N}$ and $\mathbf{m}^w \coloneq (\mathbf{m}^{w, n})_{n=1}^{N}$. Following the per-token independent noising process of diffusion forcing, we construct the noisy latent pairs as:
\begin{equation}
\begin{aligned}
    \mathbf{m}_{t_n}^{w, n} &\coloneq t_n \mathbf{m}^{w, n} + (1 - t_n)\mathbf{m}_0^{n}, \\
    \mathbf{m}_{t_n}^{l, n} &\coloneq t_n \mathbf{m}^{l, n} + (1 - t_n)\mathbf{m}_0^{n},
\end{aligned}
\end{equation}
where $n \in [1, N]$ is the frame index, $t_n \in [0, 1]$ is the $n$-th flow time, and $\mathbf{m}_0 \coloneq (\mathbf{m}_0^{n})_{n=1}^{N} \in \real^{N \times d}$ is the noise sequence. With these notations, we formulate $\loss_{DPO}$ as
\begin{equation}
\begin{aligned}
\loss_{DPO}(\theta)
&= -\mathbb{E}_{n, t_n, \mathbf{c}^n, (\mathbf{m}^{w, n},\mathbf{m}^{l, n})} \\
& \log \sigma\Big(
    - \beta \big[ \|v^{w, n}_{t_n} - v_{\theta}(\mathbf{m}^{w, n}_{t_n}, t_n, \mathbf{c}^n)\| \\
 & \quad - \|v^{w, n}_{t_n} - v_{\text{ref}}(\mathbf{m}^{w, n}_{t_n}, t_n, \mathbf{c}^n)\|  \\
 & \quad - (\|v^{l, n}_{t_n} - v_{\theta}(\mathbf{m}^{l, n}_{t_n}, t_n, \mathbf{c}^n)\| \\
 & \quad - \|v^{l, n}_{t_n} - v_{\text{ref}}(\mathbf{m}^{l,n}_{t_n}, t_n, \mathbf{c}^n)\|)\big] \Big), \\
\end{aligned}
\label{eq:dpo_train}
\end{equation}
where $\mathbf{c}^n$ is the $n$-th unified condition, $v_{\text{ref}}$ is the reference vector field model, the target vector fields for less-preferred and preferred samples are given by
\begin{equation}
v_{t_n}^{l, n} \coloneq \mathbf{m}^{l, n} - \mathbf{m}_0^{n} ~~~\text{and}~~~ v_{t_n}^{w, n} \coloneq \mathbf{m}^{w, n} - \mathbf{m}_0^{n}.
\end{equation}

\section{Experimental Details}\label{sec:apx_exp_detail}

\subsection{Inference Details}
\paragraph{Classifier-Free Guidance (CFG)} We apply independent classifier-free guidance (CFG)~\citep{instructpix2pix} for multiple driving conditions. Specifically, we compute the modified vector field $\tilde{v}_{\theta}$ as
\begin{equation}
\begin{aligned}
    \tilde{v}_\theta(x_t, t; \mathbf{c}) & \coloneq v_\theta(x_t, t; \mathbf{c}_{\{\emptyset,\emptyset\}}), \\
    & + w_{\mathbf{a}} [ v_\theta(x_t, t; \mathbf{c}_{\{\mathbf{a}, \emptyset\}}) - v_\theta(x_t, t; \mathbf{c}_{\{\emptyset,\emptyset\}})], \\
    & + w_{\mathbf{u}} [v_\theta(x_t, t; \mathbf{c}_{\{\emptyset, \mathbf{\mathbf{u}\}}}) - v_\theta(x_t, t; \mathbf{c}_{\{\emptyset, \emptyset\}}],\label{eq:apx_cfg}
\end{aligned}
\end{equation}
where $\mathbf{c}_{\{x, y\}}$ denotes a driving condition with the conditions $x$ and $y$. $\mathbf{a}$ denotes the avatar audio, and  $\mathbf{u} = \{\mathbf{a}_u, \mathbf{m}_u\}$ is the set of user audio $\mathbf{a}_u$ and user motion $\mathbf{m}_u$. $w_{\mathbf{a}}$ and $w_{\mathbf{u}}$ are the CFG scales of the avatar audio and user condition, respectively. We use 10\% dropout rate for each condition during training.

\begin{figure}[t]
\vspace{-0.15in}
\centering
\begin{minipage}{1.0\linewidth}
\begin{algorithm}[H]
\small 
\caption{\label{algo:sampling_detail} Motion inference with KV caching (Detailed)} 
\begin{algorithmic}[1]
\Require ODE timesteps $\{t_n\}^T_{n=0}$, motion generator $v_{\theta}$, video length $N$, block size $B$, lookahead size $l$, max cache size $M$, user inputs $(\mathbf{a}_{u},\mathbf{m}_{u})$, avatar audio $\mathbf{a}^i$, latent-to-frame decoder $\texttt{Dec}$, offset $\mathbf{O}$, and id latent $z_S$. 
\State \textbf{Divide} the frames into $L = \lceil N / B \rceil$ blocks
\State \textbf{Initialize} $\mathbf{KV}, \mathbf{cKV}$ $\leftarrow [~], [~]$ \hfill $\rhd$ Frame \& condition caches
\For{$i = 1$ to $L$}
    \State \textbf{Sample} Noise block $\mathbf{m}_{t_{0}}^{i} \sim \mathcal{N}(0, \mathbf{I})$, \hfill $\rhd$ 
    $\mathbf{m}_{t_{0}}^{i}\in\real^{B \times d}$
    \State \textbf{Acquire} \highlightline{User inputs 
    ($\mathbf{a}^i_{u}$, $\mathbf{m}_{u}^i$)
    and avatar audio $\mathbf{a}^i$}.
    \State \textbf{Set} $\mathbf{c}^i$ $\leftarrow$ $(\mathbf{a}^i_{u}, \mathbf{m}_{u}^i, \mathbf{a}^i)$ \hfill $\rhd$ Condition triplet
    \State \textbf{Merge offset}  \highlightlinetwo{~$\mathbf{m}^i_{t_0}$, $\mathbf{c}^i$ $\leftarrow$ \texttt{Concat}($\mathbf{m}^i_{t_0}$, $\mathbf{c}^i$; $\mathbf{O}^i$)}

    \For{$j = 0$ to $T$}
        \State \textbf{Solve ODE:} $\mathbf{m}_{t_{j+1}}^{i} \leftarrow v_{\theta}(\mathbf{m}_{t_j}^{i}, t_j; \mathbf{c}^i, \mathbf{KV}, \mathbf{cKV})$
    \EndFor
    \State \textbf{Decode} \& \textbf{Return}~~ \highlightline{$\mathbf{x}_1^i$ $\leftarrow$ $\texttt{Dec}(z_S, \mathbf{m}_1^i)$~$\in$~$\real^{B \times 3 \times H \times W}$}
    \State \textbf{Update caches}~~ $\mathbf{kv}_i$, $\mathbf{ckv}_i$ $\leftarrow$ $v_{\theta}(z_{1}^{i}, 1; \mathbf{c}^i, \mathbf{KV}, \mathbf{cKV})$    
    \If{$|\mathbf{KV}|$~$=$~$|\mathbf{cKV}|$~$=$~$M$}
    \State $\mathbf{KV}{\texttt{.pop(0)}}$~and~  $\mathbf{cKV}{\texttt{.pop(0)}}$
    \EndIf
    \State $\mathbf{KV}${\texttt{.append($\mathbf{kv}_i$)}}~and~ $\mathbf{cKV}${\texttt{.append($\mathbf{ckv}_i$)}}
    \State \textbf{Update offset} \highlightlinetwo{$\mathbf{O}^{i+1}$ $\leftarrow$ $(\mathbf{m}_1^i[-l:], \mathbf{c}^{i}[-l:])$}
\EndFor
\end{algorithmic}
\end{algorithm}
\end{minipage}
\vspace{-0.15in}
\end{figure}

\paragraph{Motion Inference Details}
We provide more details on our inference strategy in~\cref{algo:sampling_detail}.

While the lookahead attention enables to generate temporally consistent motion across the blocks, naively introducing it incurs additional latency as it requires $l$ future frames. To tackle this problem, we introduce an offset $\mathbf{O}^i$ for $i$-th block generation. Specifically, the offset $\mathbf{O}^i$ consists of the last $l$ clean motion latents and the corresponding condition from the previous block:
\begin{equation}
    \mathbf{O}^{i+1} = (\mathbf{m}_1^{i}[-l:], \mathbf{c}^{i}([-l:])),\label{eq:offset}
\end{equation}

where $i=1, \cdots, L-1$. These offset motion frames are concatenated with the current noisy motion block $\mathbf{m}^i_{t_j}\in \real^{B\times d}$ and the condition $\mathbf{c}^i \in \real^{3 \times B \times d}$ along the time axis, resulting in $l + B$ motion frames and corresponding conditions (Line 7 in~\cref{algo:sampling_detail}). Due to the flexibility of diffusion forcing, we can assign difference flow time schedules, i.e., $t=1$ for the offset and $t=t_j$ for the current noisy latent block.

\begin{figure*}[t]
    \centering
    \includegraphics[width=\linewidth]{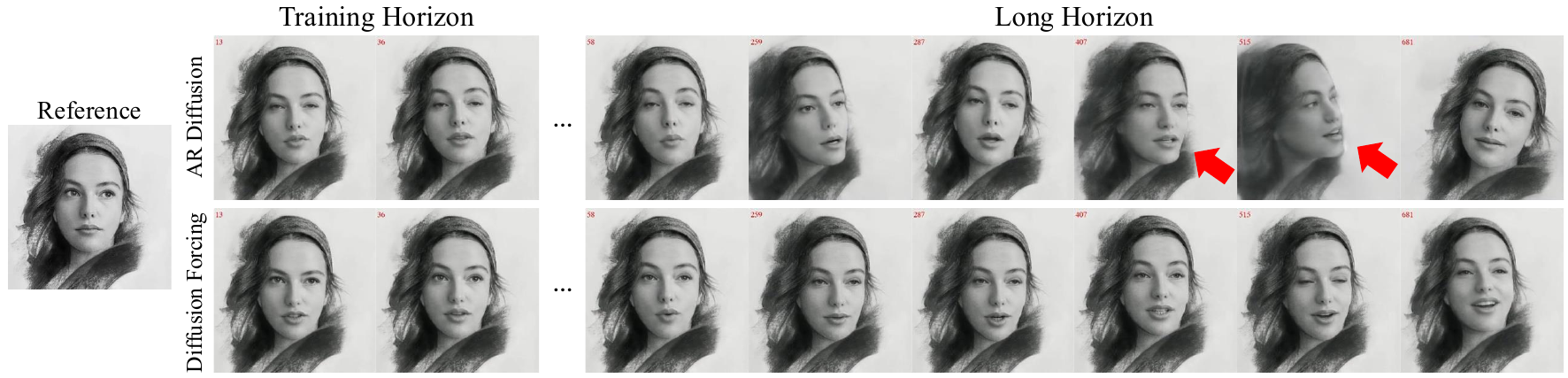}
    \vspace*{-6mm}
    \caption{Comparison of \textbf{autoregressive diffusion} and \textbf{diffusion forcing}. Autoregressive diffusion suffers from motion drift (red arrow) over the long horizon, whereas diffusion forcing maintains stable motion generation over the long horizon.}\label{fig:apx_ablation_ar_diffusion}
    \vspace*{-2mm}
\end{figure*}

As we compute the modified vector field $\tilde{v}_\theta$ for CFG as in Eq.~\eqref{eq:apx_cfg}, we separately cache and update the KV of these vector fields $v_\theta (\cdot; \mathbf{c}_{\{\emptyset, \emptyset\}})$, $v_\theta (\cdot; \mathbf{c}_{\{\mathbf{a}, \emptyset\}})$, and $v_\theta (\cdot; \mathbf{c}_{\{\emptyset, \mathbf{u}\}})$. To obtain the KV caches of the clean block, we compute all three by reusing the generated motion block along with the existing KV caches~\citep{self_forcing}. Moreover, due to the introduction of lookahead attention and offset, the KV caches are updated except the for last $l$ frames and these frames are provided by the offset. Therefore, the maximum cache size is $M = L - B - l = 38$.
\begin{figure*}[h]
\captionof{table}{
\textbf{Ablation study with additional metrics} on user motion and preference optimization. 
``w/ $\mathbf{m}_u$'' indicates whether the user motion latent $\mathbf{m}_u$ is provided as input to the model during both training and inference.
} \label{tab:apx_ablation_realtalk}
\vspace{-0.0in}
\centering
    \vspace*{-2mm}
    \resizebox{0.92\textwidth}{!}{
        \renewcommand{\arraystretch}{1.0}
        \renewcommand{\tabcolsep}{6pt}
        \begin{tabular}{c c c c c c c c c c c}
        \toprule
         \multicolumn{2}{c}{Method} & \multicolumn{2}{c}{Reactiveness} & \multicolumn{2}{c}{Motion Richness} & \multicolumn{3}{c}{Visual Quality}  & \multicolumn{2}{c}{Lip Synchronization} \\
         \cmidrule(l{2pt}r{2pt}){1-2}
        \cmidrule(l{2pt}r{2pt}){3-4}
        \cmidrule(l{2pt}r{2pt}){5-6}
        \cmidrule(l{2pt}r{2pt}){7-9}
        \cmidrule(l{2pt}r{2pt}){10-11}
        w/ $\mathbf{m}_u$ & DPO & rPCC-Exp $\downarrow$ & rPCC-Pose $\downarrow$ & SID $\uparrow$ & Var $\uparrow$ & FID $\downarrow$ & FVD $\downarrow$ & CSIM $\uparrow$ & LSE-D $\downarrow$ & LSE-C $\uparrow$\\
        
        \midrule        
        \redx & \redx       & 0.052 & 0.175 & 2.165 & 1.586 & 28.746 & 185.593 & 0.818 & 8.260 & 6.423\\
        \greencheck & \redx               & 0.042 & 0.146 & 2.236 & 1.408 & 25.600 & 175.322 & \textbf{0.854} & 8.160 & \textbf{6.803} \\
        \midrule        
        \greencheck & \greencheck     & \textbf{0.003} & \textbf{0.036} & \textbf{2.442} & \textbf{1.734} & \textbf{24.328} & \textbf{170.874} & 0.833  &  \textbf{8.060} & 6.723\\
        \bottomrule
        \end{tabular}
        }
\vspace{-0.1in}
\end{figure*}

\subsection{Training Details}
We train the vector field model $v_{\theta}$ in Eq.~\eqref{eq:train_ours} for 2000k steps while freezing the motion latent auto-encoder. We use L1 distance for $\| \cdot \|$. For fine-tuning the model using the proposed preference optimization method in Eq.~\eqref{eq:total_objective}, we set the balancing coefficient to $\lambda=0.1$ and the deviation parameter to $\beta=1000$. We initialize the reference model $v_{\text{ref}}$ with the same weights as the trained $v_{\theta}$. We fine-tune $v_\theta$ for 5k steps and observe that additional tuning does not yield further performance gains. 

\subsection{Baseline Implementation}
One major challenge of evaluating an interactive head avatar model is the \textbf{absence of official implementation of baseline methods}. To bridge this gap, we reproduce INFP~\citep{infp} on the motion latent space of ~\citep{float}, following its core module, Motion Guider and denote the reproduced model as INFP*. Based on the description in the original paper, we adopt a bidirectional Transformer encoder for motion generation, where a single window consists of $N=75$ frames and additional $10$ frames serve as context frames. For Motion Guider, we set $K=64$ for both verbal and non-verbal motion memory banks. We train INFP* on our dataset for 2000k steps.

\subsection{Additional Ablation Studies}\label{sec:apx_additional_ablation}
\paragraph{Comparison with Autoregressive Diffusion} We compare our motion latent diffusion forcing with standard autoregressive diffusion, where the motion generator is conditioned on clean context motion latents. Specifically, we train the motion generator in an autoregressive diffusion manner using four clean context blocks (40 frames) and one noisy block (10 frames). As shown in~\cref{fig:apx_ablation_ar_diffusion}, autoregressive diffusion suffers from degraded long-horizon generation, whereas diffusion forcing is much more robust to motion drift, highlighting its necessity.

\paragraph{Ablation on Motion Motion block Size} In~\cref{tab:apx_ablation_block_size}, we provide an ablation study on motion block size under a fixed number of training frames $N=50$. Increasing block size (i.e., reducing the number of frames in each block) leads to lower latency while achieving quantitative performance. Conversely, reducing the block size (i.e., increasing the number of frames in each block) can improve the temporal consistency (FVD) and Lip-sync quality (LSE-D) with higher latency.

\begin{figure}[t]
    \centering
    \includegraphics[width=1\linewidth]{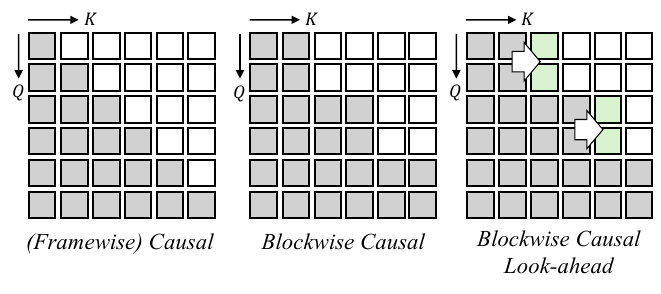}
    \vspace{-0.3in}
    \caption{\textbf{Attention Mask Comparison.} (Left) framewise causal mask; (Middle) blockwise causal mask; (Right) blockwise causal look-ahead mask (Ours).} \label{fig:mask_compare}
    \vspace{-0.1in}
\end{figure}

\paragraph{Additional Quantitative Results} In~\cref{tab:apx_ablation_realtalk}, we present the ablation studies on our model with additional metrics, including Visual Quality and Lip Synchronization. We provide a video results of the ablation study. Please refer to the videos \textit{``02\_ablation\_wo\_user\_motion.mp4''} and \textit{``02\_ablation\_wo\_DPO.mp4''}. Moreover, we provide video ablation results on the attention mask, where each masking method is illustrated in~\cref{fig:mask_compare}. The motion jittering observed when using only the blockwise causal mask is clearly visible in the video results, yet difficult to capture with quantitative metrics. We highly recommend watching the ablation video \textit{``02\_ablation\_attention\_mask\_XX.mp4''}.

\subsection{Evaluation Metrics}
\paragraph{Reactiveness and Motion Richness} Reactiveness and Motion Richness are computed using the EMOCA-based~\citep{emoca} 3D morphable models (3DMMs) that model the facial dynamics via 50-dim expression parameters and 6-dim pose parameters. We extract those parameters using an off-the-shelf 3DMM extractor, SPECTRE~\citep{spectre}, for each video frame. Let us denote $x\in\real^{N\times d}$ as the ground-truth user parameters, $y\in\real^{N\times d}$ as the ground-truth avatar parameters, and $\hat{y}\in\real^{N\times d}$ as the generated avatar parameters. $L$ is the number of frames and $d$ is the feature dimension. As reported in~\cref{tab:quantitative_realtalk} and~\cref{tab:compare_listening_vico}, we can compute rPCC, SID, Var, and FD for expression and pose, respectively.

\begin{itemize}[leftmargin=1.5em]
\item \textbf{rPCC} (residual Pearson Correlation Coefficients)~\citep{dim} is to measure the motion synchronization between the user parameters and avatar parameters. Specifically, L1 distance is used to measure the discrepancy between generated PCC and ground-truth PCC where we define PCC as a function of $z\in\real^{N\times d}$ given ground-truth user parameters $x$:
\begin{equation}
    \text{PCC}(z|x) = \frac{\sum (z_i - \bar{z})(x_i - \bar{x})}{\sqrt{\sum(z_i - \bar{z})^2 \sum (x_i - \bar{x})^2}},
\end{equation}
where $i\in[0,N]$ is the frame index, and $\bar{z}$ and $\bar{x}$ denote the mean of $z$ and $y$, respectively. Based on this notation, we can define rPCC as $|\text{PCC}(y|x) - \text{PCC}(\hat{y}|x)|$. 

\item \textbf{SID} (Shannon Index for Diversity)~\citep{l2l} is to measure the motion diversity of the generated avatars using $K$-means clustering on 3DMM parameters. Following~\citep{l2l}, we compute the average entropy (Shannon index) of the clusters with $K=15, 9$ for expression and pose, respectively.

\item \textbf{Var} is the variance of the parameters from generated avatars, which is computed along the time axis and then averaged along the feature axis. 

\item \textbf{FD} (Frechet Distance)~\citep{fid} measures the distance between the expression and pose distributions of the generated avatars and the ground truth by calculating
\begin{equation}
    |\mu_{\hat{y}} - \mu_{y}| + \text{tr}(\Sigma_{\hat{y}} + \Sigma_{y} - 2 (\Sigma_{\hat{y}} \Sigma_{y})^{\frac{1}{2}}), \label{eq:fd}
\end{equation}
where $\mu$ and $\Sigma$ are the mean and the covariance matrix, respectively.
\end{itemize}

\begin{table}
\centering
\caption{Ablation studies on motion block sizes on RealTalk.}\label{tab:apx_ablation_block_size}
\vspace{-0.1in}
\resizebox{0.99\linewidth}{!}{
\begin{tabular}{l|cccccc}
\toprule
\# blocks & Latency (s) $\downarrow$ & rPCC-Exp $\downarrow$ & rPCC-Pose $\downarrow$ & FVD $\downarrow$ & SID $\uparrow$ & LSE-D $\downarrow$ \\
\midrule
10 & \textbf{0.3} & 0.012 & 0.056 & 222.47 & \underline{2.355} & 7.290\\
2  & 1.5 & \textbf{0.003} & \textbf{0.031} & \textbf{155.81} & 2.145 & \textbf{6.555}\\
\midrule
\textbf{5 (Ours)}  & 0.5 & \textbf{0.003} & \underline{0.036} & \underline{170.87} & \textbf{2.442} & \underline{6.723} \\
\bottomrule
\end{tabular}
}
\vspace{-0.1in}
\end{table}

\paragraph{Visual Quality} We utilize FID~\citep{fid} and FVD~\citep{fvd} to assess the image and video quality of the generated avatars, and CSIM~\citep{arcface} to measure the identity preservation performance of avatar generation models.
\begin{itemize}[leftmargin=1.5em]
\item \textbf{FID} (Frechet Inception Distance) measures the quality of the generated frames by comparing the distribution of image features extracted from a pre-trained feature extractor~\citep{fid}. The FD computation in Eq.~\eqref{eq:fd} is adopted using the extracted image features.
\item \textbf{FVD} (Frechet Video Distance) quantifies the spatio-temporal quality of the generated videos by comparing the feature distributions of real and generated videos in a learned video feature space~\citep{fvd}. It reflects both frame-wise quality and temporal consistency. The FD computation in Eq.~\eqref{eq:fd} is adopted using the extracted video features.
\item \textbf{CSIM} (Cosine Similarity for Identity Embedding) evaluates identity preservation by computing cosine similarity between the facial embeddings from the generated and the source image, extracted using ArcFace~\citep{arcface}.
\end{itemize}

\paragraph{Lip Synchronization} We compute LSE-D and LSE-C~\citep{syncnet} to assess the alignment between the generated lip motion and the corresponding audio.
\begin{itemize}[leftmargin=1.5em]
\item \textbf{LSE-D} and \textbf{LSE-C} (Lip Sync Error Distance and Confidence):
Both metrics are derived from a pre-trained SyncNet-based audio–visual synchronization model. LSE-D measures the distance between the audio and lip embeddings, where lower values indicate better synchronization. LSE-C measures the confidence score of synchronized audio–visual pairs, where higher values indicate more accurate lip-audio alignment.
\end{itemize}

\subsection{Human Evaluation} 
\begin{figure*}[t]
    \centering
    \includegraphics[width=\linewidth]{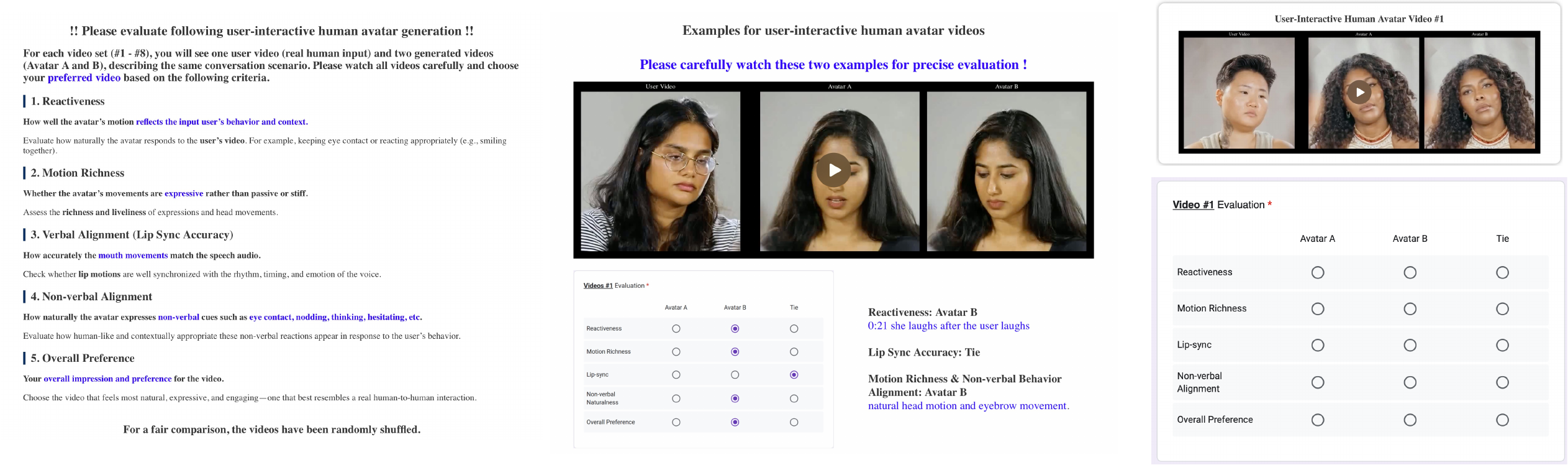}
    \caption{\textbf{Human evaluation interface}. (Left) Instructions for human evaluation; (Middle) A reference sheet for consistent evaluation; (Right) Test and answer sheet.}\label{fig:apx_human_eval}
\end{figure*}

In~\cref{fig:apx_human_eval}, we show the interface used for our human evaluation. To improve the evaluation consistency, we additionally provided participants with a reference test and answer sheet. We asked 42 participants to compare 12 videos based on 5 evaluation metrics and indicate their preference. We also provide a video test sheet. Please refer to \textit{``04\_human\_evaluation\_XX.mp4''}.

%%%%%%%%%%%%%%%%%%%%%%%%%%%%%%%%%%%%%%%%%%%%%%%%%%%%%%%%%%%%%%%%%%%%%%%%%%%%%%%%%%%%%%%%%%%%%%%%
%%%%%%%%%%%%%%%%%%%%%%%%%%%%%%%%%%%%%%%%%%%%%%%%%%%%%%%%%%%%%%%%%%%%%%%%%%%%%%%%%%%%%%%%%%%%%%%%

\subsection{Supplementary Visual Results}
\paragraph{Comparison with Interactive Head Avatar}
We provide the video results to further support the visual results in~\cref{fig:sota_compare}. Please refer to \textit{``01\_interactive\_avatar\_comparison\_XX.mp4''}. We also provide a video comparison results using the DEMO videos of Official INFP~\citep{infp}. Please refer to \textit{``01\_interactive\_avatar\_comparison\_demo.mp4''}

\begin{figure*}[t]
    \centering
    \includegraphics[width=0.99\linewidth]{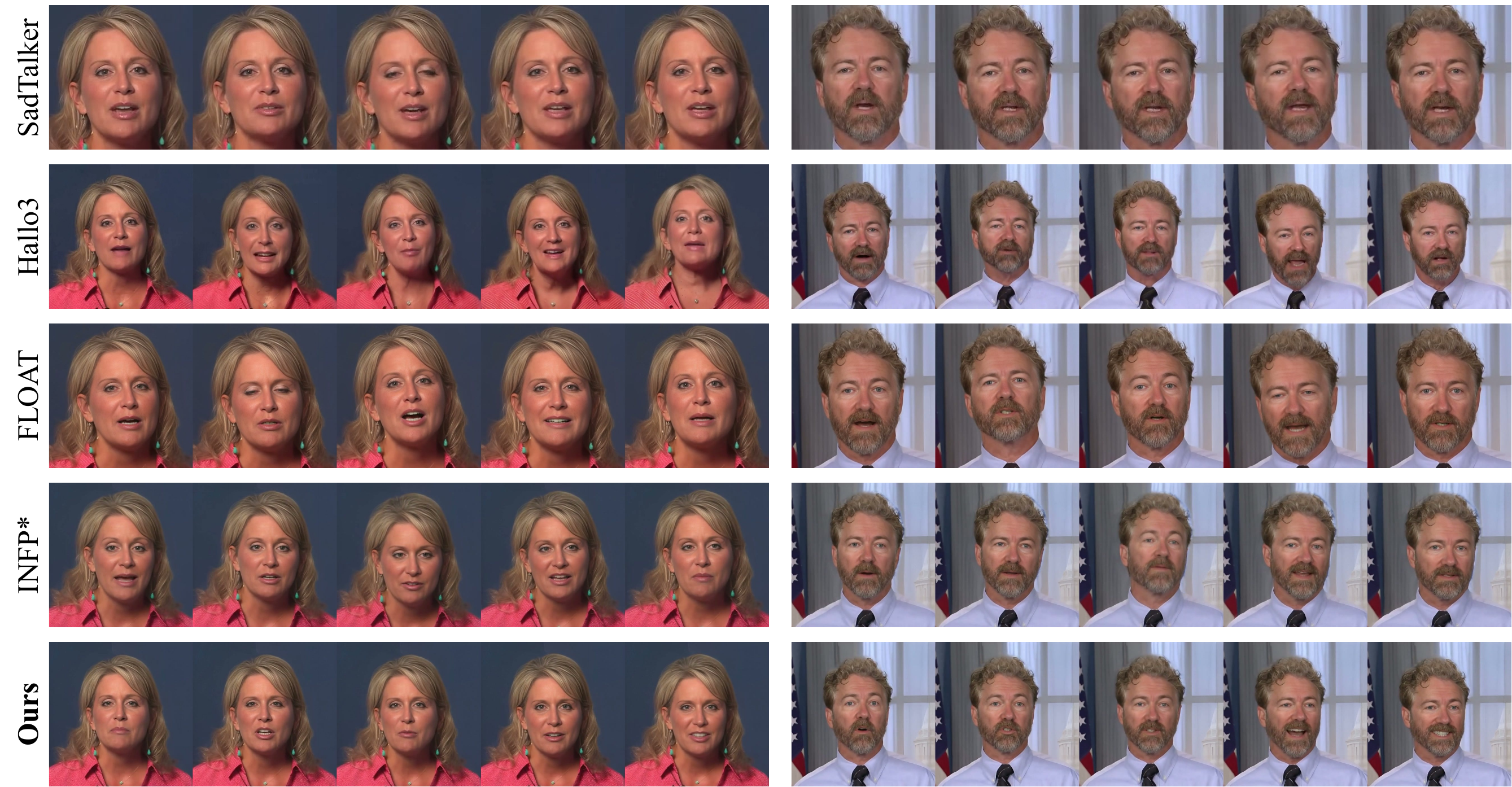}
    \caption{\textbf{Qualitative comparison on talking head avatar generation}.\label{fig:apx_talking_head}}
\end{figure*}

\paragraph{Comparison with Talking Head Avatar}
In~\cref{fig:apx_talking_head}, we compare our model with SadTalker~\citep{sadtalker}, Hallo3~\citep{hallo3}, FLOAT~\citep{float}, and INFP*~\citep{infp} for talking head avatar generation bu dropping the user condition at inference. Avatar Forcing can generate competitive results compared to state-of-the-art models, while our model successfully reflects user signals. We also provide the video comparison results. Please refer to \textit{``03\_talking\_XX.mp4''}.

\begin{figure*}[t]
    \centering
    \includegraphics[width=0.99\linewidth]{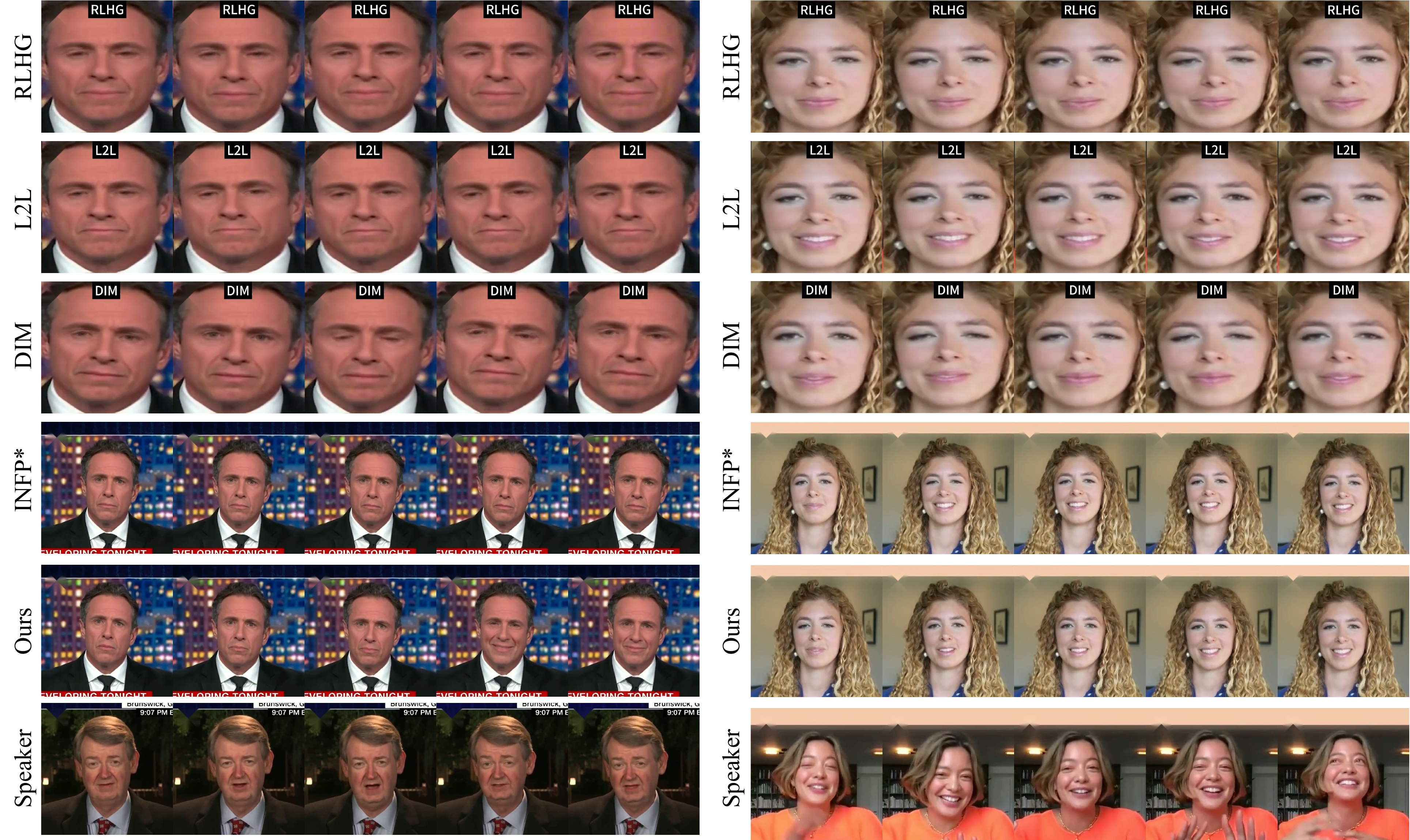}
    \caption{\textbf{Qualitative comparison on listening head avatar generation}.\label{fig:apx_listening_head}}
\end{figure*}

\paragraph{Comparison with Listening Head Avatar}
In~\cref{fig:apx_listening_head}, we compare our model with RLHG~\citep{rlhg}, L2L~\citep{l2l}, DIM~\citep{dim}, and INFP*~\citep{infp} for listening head avatar generation. Avatar Forcing can generate competitive results with more expressive facial expression. Please refer to \textit{``03\_listening\_XX.mp4''} for video results.

\section{Discussion}\label{sec:apx_discussion}
\paragraph{Ethical Consideration} Our method can generate more engaging and interactive head-avatar videos, broadening positive applications such as virtual avatar chat, virtual education, and other communication tools by providing users with a more immersive experience. However, realistic interactive head avatar videos also pose risks of misuse, including identity spoofing or malicious deepfakes. Adding watermarks to generated videos
and applying a restricted license can help mitigate these risks. We also encourage the community to use our generated data to train deepfake detection models.

\paragraph{Limitation and Future Work}
Our system focuses on modeling interactive conversations through a head-motion latent space, which enables natural and expressive interactions. This design limits the modeling of richer bodily cues, such as hand gestures, that contribute to more dynamic communication. Moreover, while our model captures user-driven conversational cues via motion latents, certain scenarios may require more explicit controllability, such as directing eye gaze or emphasizing emotional shifts. We believe that incorporating additional user signals, including eye-tracking or emotion-tracking inputs, can address these limitations. Since our framework imposes no architectural constraints on adding new conditions, such signals can be incorporated in future extensions of our system. While diffusion forcing is robust for long-horizon generation, it does not fully address exposure bias. Addressing this issue in the motion latent space remains future work.

%%%%%%%%%%%%%%%%%%%%%%%%%%%%%%%%%%%%%%%%%%%%%%%%%%%%%%%%%%%%%%%%%%%%%%%%%%%%%%%%%%%%%%%%%%%%%%%%
%%%%%%%%%%%%%%%%%%%%%%%%%%%%%%%%%%%%%%%%%%%%%%%%%%%%%%%%%%%%%%%%%%%%%%%%%%%%%%%%%%%%%%%%%%%%%%%%

\newpage

%%%%%%%%%%%%%%%%%%%%%%%%%%%%%%%%%%%%%%%%%%%%%%%%%%%%%%%%%%%%%%%%%%%%%%%%%%%%%%%%%%%%%%%%%%%%%%%%
%%%%%%%%%%%%%%%%%%%%%%%%%%%%%%%%%%%%%%%%%%%%%%%%%%%%%%%%%%%%%%%%%%%%%%%%%%%%%%%%%%%%%%%%%%%%%%%%

\FloatBarrier
\clearpage


\end{document}